\documentclass[runningheads]{llncs}

\PassOptionsToPackage{table}{xcolor}

\usepackage{accv}

\usepackage{accvabbrv}

\usepackage{graphicx}
\usepackage{booktabs}
\usepackage{multirow}
\usepackage{placeins}  
\usepackage{pifont}  
\usepackage{colortbl}  
\usepackage{tikz}
\usetikzlibrary{positioning, arrows.meta, calc, decorations.pathreplacing, fit, backgrounds}
\usepackage{amssymb}
\usepackage{bm}

\usepackage[accsupp]{axessibility}  

\usepackage{hyperref}
\hypersetup{hidelinks, pdftitle={SCCM: Spherically Consistent Coarse Matching for ERP Dense Feature Correspondence},
  pdfauthor={Gyeonggwan Lee, Eunsoo Im, Seunghwan Hong, Junghun Suh},
  pdfsubject={ACCV 2026},
  pdfkeywords={Dense matching, 360-degree images, Spherical geometry}}

\begin{document}

\title{\texorpdfstring{%
SCCM: Spherically Consistent Coarse Matching for ERP Dense Feature Correspondence%
}{SCCM: Spherically Consistent Coarse Matching for ERP Dense Feature Correspondence}}

\titlerunning{SCCM: Spherically Consistent Coarse Matching for ERP}

\author{Gyeonggwan Lee\inst{1,2} \and
        Eunsoo Im\inst{1} \and
        Seunghwan Hong\inst{1} \and
        Junghun Suh\inst{1}}

\authorrunning{G.~Lee et al.}

\institute{Kakao Mobility Corp., Seongnam, Republic of Korea \and
Korea University, Seoul, Republic of Korea\\
\email{\{gandan.lee,soo.hd,logan.sh,jude.suh\}@kakaomobility.com}\\
Project page: \url{https://gandanlee.github.io/sccm/}}

\maketitle

\begin{abstract}
Dense feature matching between 360$^\circ$ panoramas underpins omnidirectional
pose estimation, 3D reconstruction, and SLAM. Such panoramas are stored in the
equirectangular projection (ERP), which unrolls the viewing sphere onto a flat
chart and thereby introduces three distinct distortions---a longitudinal seam
(topology), latitude-dependent stretch (metric), and non-uniform pixel area
(area)---that the coarse stage of perspective-trained dense matchers does not
model, so these matchers degrade systematically on ERP.

We show that correcting the three distortions at the coarse-stage
interfaces where they arise---pairwise distortions in attention,
per-pixel distortion in covisibility gating---improves PCK@$1^\circ$
from $0.230$ to $0.275$ on Matterport3D under a fixed coarse scaffold,
with the refiner architecture unchanged---our central result.
Concretely, SCCM (Spherically Consistent Coarse Matching) augments a
chart-naïve cross-attention/dual-softmax coarse matcher with two
sphere-derived priors: Spherical Positional Attention (SPA) pairs a
yaw-periodic RoPE (topology) with a tangent-plane bias
(metric), and Area-Aware Covisibility (AAC) applies a pre-sigmoid
log-area correction (area).
The chart-naïve scaffold serves as a controlled reference, separating
the scaffold-replacement effect from the spherical-prior effect.
Instantiated in the RoMa~V1 framework with the same frozen encoder, refiner
architecture, and loss, SCCM also outperforms the ERP-native EDM ($0.163$) and an
ERP-retrained RoMa~V1 ($0.198$) under a unified ERP dense matching
protocol, while perspective-trained matchers largely fail on
ERP. It further transfers zero-shot to Stanford2D3D and, when trained on
outdoor Holo360D, leads there as well.

\end{abstract}

\section{Introduction}
\label{sec:intro}

\begin{figure}[!t]
\centering
\includegraphics[width=\textwidth]{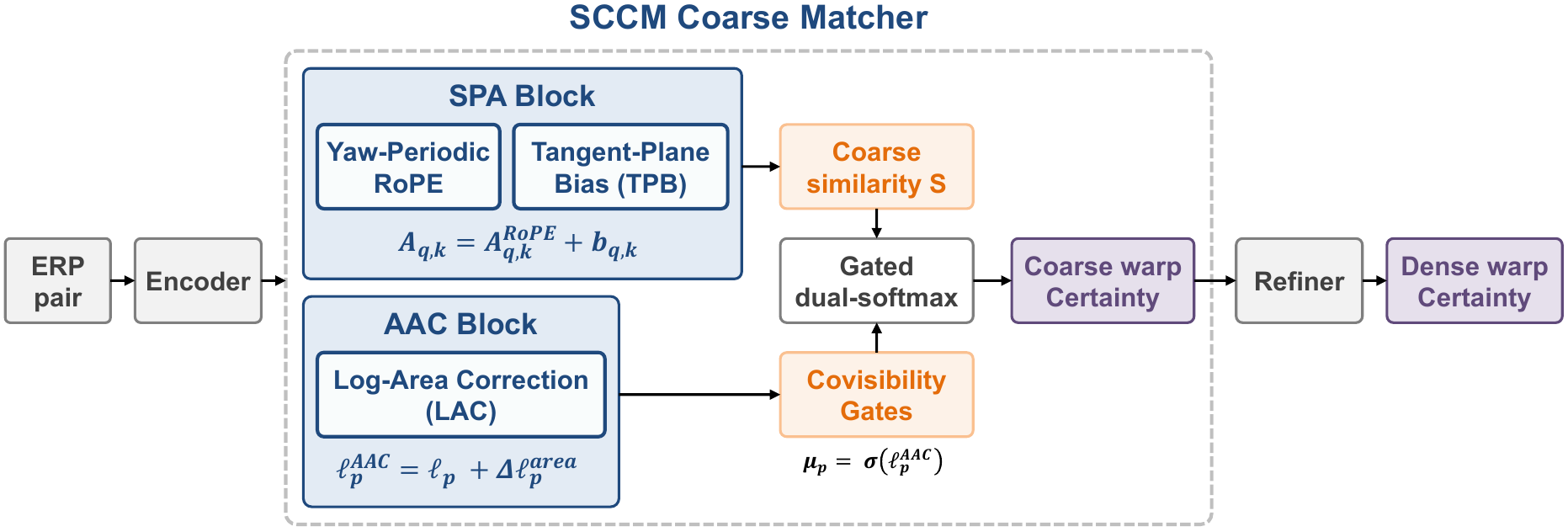}
\caption{\textbf{SCCM coarse matcher: pipeline overview.}
SCCM injects sphere-derived priors into a fixed chart-naïve coarse scaffold
(denoted R1 in Sec.~\ref{sec:method}), while keeping the encoder, refiner
architecture, and loss unchanged. The inherited
scaffold consists of cross-/self-attention, covisibility gating, and gated
dual-softmax. The \emph{blue} modules are the proposed sphere-aware priors:
yaw-periodic RoPE (topology) and Tangent-Plane Bias (metric) in
SPA (Fig.~\ref{fig:spa_module}), and Log-Area Correction (LAC; area) in AAC
(Fig.~\ref{fig:aac_module}); \emph{orange} boxes denote coarse matching signals
passed to gated dual-softmax, \emph{grey} boxes are inherited unchanged,
and \emph{purple} boxes are outputs. The output is a dense warp $W_{A\to B}$
with per-pixel certainty $c$.}
\label{fig:teaser}
\end{figure}


A 360$^\circ$ camera captures the full viewing sphere in a single shot, 
so dense correspondences between two panoramas---a matching location for every pixel---
provide wide-overlap constraints 
for relative pose, 3D reconstruction, and omnidirectional SLAM. 
Panoramas are almost always stored in the equirectangular projection (ERP), 
the chart that unrolls the sphere onto a flat rectangle, 
so in practice dense matching on 360$^\circ$ content operates on ERP images. 
Modern dense and semi-dense matchers~\cite{dkm,roma,loftr} solve the task in two stages:
a coarse matcher first proposes low-resolution match anchors, 
and a refiner then sharpens them into precise correspondences. 
These matchers achieve strong results on perspective images 
but degrade systematically on ERP, 
where the chart's latitude-dependent stretch and longitudinal seam 
violate the planar assumptions built into their coarse stage.

ERP distortion is not a single artifact but a composite of three
computational challenges.
\textbf{Topology}: the longitudinal $\pm\pi$ seam breaks positional continuity.
\textbf{Pairwise metric}: a longitudinal stretch by $1/\cos\varphi$ grows toward
the poles (here $\varphi$ is latitude, $0$ at the equator and
$\pm\tfrac{\pi}{2}$ at the poles, so $1/\cos\varphi\!\to\!\infty$), making
Euclidean chart offsets poorly aligned with true geodesic offsets, especially
near the poles.
\textbf{Per-pixel area}: the $\cos\varphi$ scaling of the sphere's area
element (the patch of sphere a single pixel covers, which shrinks toward the
poles) over-represents polar regions in pixel-uniform sampling and can
bias per-pixel matchability estimation toward polar candidates.
The first two are \emph{pairwise}---they concern the relation between two
feature locations, so we correct them inside \emph{attention} (which scores
pairwise interactions between feature locations); the third is \emph{per-pixel}---it concerns
whether a single pixel is worth matching at all, so we correct it inside the
covisibility \emph{gate} (a per-pixel keep/suppress weight). A single post-hoc
adjustment of the final matching cost risks conflating these distinct pairwise
and per-pixel effects; we therefore inject each correction at the interface where
its distortion arises.

Concretely, we first define a chart-naïve cross-/self-attention $+$
dual-softmax coarse scaffold, then introduce \textbf{SCCM} (Spherically
Consistent Coarse Matching) as the \emph{same} scaffold augmented with two
sphere-aware blocks, SPA and AAC, while keeping the encoder, refiner
architecture, and loss unchanged (Fig.~\ref{fig:teaser}). SCCM targets the
coarse stage, where ERP chart distortions first corrupt the anchors passed to
refinement.

The \textbf{Spherical Positional Attention (SPA)} block augments
the attention layers of the coarse cascade with 
a yaw-periodic Rotary Position Embedding (RoPE~\cite{roformer}; topology) 
and a tangent-plane bias on the sphere (pairwise metric), 
built on the Continuous Position Bias (CPB~\cite{swinv2}) framework.

The \textbf{Area-Aware Covisibility (AAC)} block augments
the covisibility
gate with a pre-sigmoid log-area correction (per-pixel area).
Both blocks follow the distortion-to-interface principle: pairwise geometry
enters attention, and per-pixel area enters covisibility gating.

Importantly, part of SCCM's full-system gain comes from the scaffold
replacement itself, not from spherical geometry. We therefore use the
chart-naïve scaffold as a controlled reference and report the
scaffold-replacement and spherical-prior components separately
(Sec.~\ref{sec:exp-comparison}).

\noindent In summary, we make three contributions:
\begin{itemize}
\item \textbf{Distortion-to-interface formulation.} We formulate ERP coarse
      matching as a distortion-to-interface problem: topology and metric
      distortions are pairwise and are corrected inside attention, while area
      distortion is per-pixel and is corrected inside covisibility gating.
\item \textbf{SPA and AAC.} A spherical attention prior (SPA) combining
      yaw-periodic RoPE for seam-continuous topology with a tangent-plane bias
      for geodesic pairwise-metric calibration, and an area-aware covisibility
      prior (AAC) applying a pre-sigmoid log-area correction to the gate.
\item \textbf{Controlled evaluation protocol.} An evaluation that separates
      scaffold gains from spherical-prior gains, with the priors alone improving
      PCK@$1^\circ$ from $0.230$ to $0.275$ under a fixed scaffold.
\end{itemize}


\section{Related Work}
\label{sec:related}

\noindent\textbf{Perspective and ERP matchers.}
Perspective matchers---SuperGlue~\cite{superglue}, LoFTR~\cite{loftr},
DKM~\cite{dkm}, and the RoMa family~\cite{roma,romav2}---do not explicitly model
ERP chart topology or latitude-dependent metric distortion in their coarse
stage, which can leave ERP-specific coarse errors for the refiner to resolve.

3D pointmap regressors (DUSt3R~\cite{dust3r}, MASt3R~\cite{mast3r}, VGGT~\cite{vggt})
also yield dense correspondences as a by-product of geometry prediction, 
but are likewise trained on perspective views; 
evaluated zero-shot on ERP, MASt3R and VGGT fall below the ERP-native EDM (Supp.~Sec.~Q).

Sparse ERP matchers such as SPHORB~\cite{sphorb} and
SphereGlue~\cite{sphereglue} apply geometry at isolated keypoints, whereas
dense matching exposes \emph{every} pairwise matching score to chart distortion.
CoMatch~\cite{comatch} introduces covisibility-aware semi-dense matching for
perspective images (covisibility-guided token condensing and attention); SCCM
instead uses covisibility as a sphere-area-corrected gate for ERP
(Sec.~\ref{sec:method-aac}).
EDM~\cite{edm} is the closest prior work: an ERP-native dense matcher with
spherical input embeddings and sphere-aware refinement. The difference is one of \emph{where}
geometry is injected: EDM injects spherical information at the input and
refinement levels, whereas SCCM injects geometry directly into the two generic coarse-stage
decision interfaces---the pairwise attention logits and the per-pixel
covisibility logits (Sec.~\ref{sec:method})---which EDM does not explicitly
correct in the same way.

\noindent\textbf{Sphere-aware encodings.}

Spherical convolutions~\cite{sphericalcnn,spherenet,equiconv}, tangent
remapping~\cite{tangent_images}, and panoramic transformers~\cite{panoswin,sphereuformer}
encode geometry within a \emph{single} image and largely leave the relative
structure between two feature grids unaddressed, while standard planar RoPE
variants used in vision~\cite{roformer,rope2d} encode 2D chart offsets but do
not enforce the $2\pi$ longitudinal periodicity required at the ERP seam.
Recent panoramic models handle the seam or chart distortion in their
positional encodings or tokens: Dense360~\cite{dense360} folds horizontal
indices symmetrically about the image center and rescales them by one global
latitude factor; RoPE Rolling~\cite{panosplatt3r,eagle360} shifts the seam to a
different longitude per head rather than removing it; SpheRoPE~\cite{spherope}
snaps high-frequency harmonics to integers and encodes low frequencies on the
sphere, for generation; PanoFormer~\cite{panoformer} uses tangent-patch tokens
for monocular depth. These encodings support panoramic understanding,
generation, depth, or two-view reconstruction; SCCM addresses dense angular
correspondence by placing seam-periodic and latitude-aware priors directly
in coarse attention and covisibility gating.

\section{Preliminaries}
\label{sec:prelim}

\noindent\textbf{ERP geometry.}
An ERP pixel at normalized coordinates $(u,v)\in[-1,1]^2$ maps to
longitude/latitude $(\lambda,\varphi)=(\pi u,\,\tfrac{\pi}{2}v)$ and thence to a
unit viewing ray on the sphere $S^2$,
\begin{equation}
\mathbf{r}(\varphi,\lambda)=
(\cos\varphi\sin\lambda,\;\sin\varphi,\;\cos\varphi\cos\lambda)\in S^2 ,
\label{eq:ray}
\end{equation}
whose area element is $\cos\varphi$. All SCCM priors derive from this
map.

We briefly recall the three standard mechanisms that SCCM modifies:
RoPE for relative phase, CPB for additive attention bias, and covisibility
gating for filtering non-matchable pixels. This section only fixes notation;
the sphere-specific modifications are introduced in Sec.~\ref{sec:method}.
\begin{enumerate}
\item \textbf{RoPE.} \emph{Rotary Position Embedding}~\cite{roformer} rotates the
query/key vectors by position-dependent phases so that their dot product carries
relative-position information.
SPA makes the longitudinal phase yaw-periodic across the seam (Sec.~\ref{sec:method-spa}).
\item \textbf{CPB.} \emph{Continuous Position Bias}~\cite{swinv2}
maps a continuous relative coordinate through a small shared MLP to a scalar
added to the pre-softmax attention logits. SPA feeds it a spherical log-map offset
instead of a planar pixel offset (Sec.~\ref{sec:method-spa}).
\item \textbf{Covisibility gating.} Occlusion and limited overlap leave many
pixels without a valid match, so dense matchers predict a per-pixel matchability
logit $\ell$ and gate $\mu=\sigma(\ell)\in[0,1]$ to down-weight non-covisible
pixels before \emph{dual-softmax} matching---related notions include CoMatch's
covisibility~\cite{comatch}, LoFTR's \emph{confidence}~\cite{loftr}, and RoMa's
\emph{certainty}~\cite{roma}. AAC modifies this pre-sigmoid logit with a
sphere-area correction (Sec.~\ref{sec:method-aac}).
\end{enumerate}

\section{Method}
\label{sec:method}

We first define a chart-naïve coarse-matching scaffold, denoted \textbf{R1}:
cross-/self-attention~\cite{superglue} followed by covisibility-gated
dual-softmax~\cite{loftr,matchformer} (Sec.~\ref{sec:prelim}). R1 treats ERP as
a flat image---no spherical positional encoding, no tangent-plane metric bias,
and no area correction. R1 serves two roles: a strong chart-naïve baseline in
its own right (Sec.~\ref{sec:exp-comparison}), and the fixed scaffold on which
spherical priors are isolated (Sec.~\ref{sec:exp-ablation}). \textbf{SCCM} is then R1 augmented with
SPA (Sec.~\ref{sec:method-spa}) and AAC (Sec.~\ref{sec:method-aac}).

We instantiate this scaffold inside the RoMa V1~\cite{roma} framework 
because its coarse stage is cleanly separable and its training recipe is reproducible.
RoMa V2~\cite{romav2}, in contrast, couples its coarse stage with other architectural changes 
and lacked a comparable public training recipe at the time of writing, 
so an ablation on it would be confounded (Supp.~Sec.~G); 
it therefore appears only as a zero-shot perspective baseline. 
Concretely, we replace RoMa V1's Gaussian-process coarse stage with R1, 
so that R1, SCCM, and every intermediate ablation row occupy the same slot. 
In all variants we keep the DINOv2-Large~\cite{dinov2} encoder, 
the ConvRefiner architecture, and the loss unchanged, freeze the encoder, 
and train all remaining modules---including the refiner---from random initialization.

Both modules act only at generic coarse-stage interfaces: SPA modifies the
attention logits, and AAC the covisibility logits. The additions are
lightweight---a ${\sim}1.1$K-parameter shared MLP for TPB and one scalar per
covisibility side for Log-Area Correction (LAC), with a cached inference overhead of about $5\%$
(Supp.\ Sec.~L)---and each is added with a geometry-motivated initialization (TPB
zero-initialized as a no-op, LAC at $\alpha_{\mathrm{LAC}}\!=\!1$), keeping the
cumulative ablation strictly comparable.

\begin{figure}[!ht]
\centering
\includegraphics[width=0.95\textwidth]{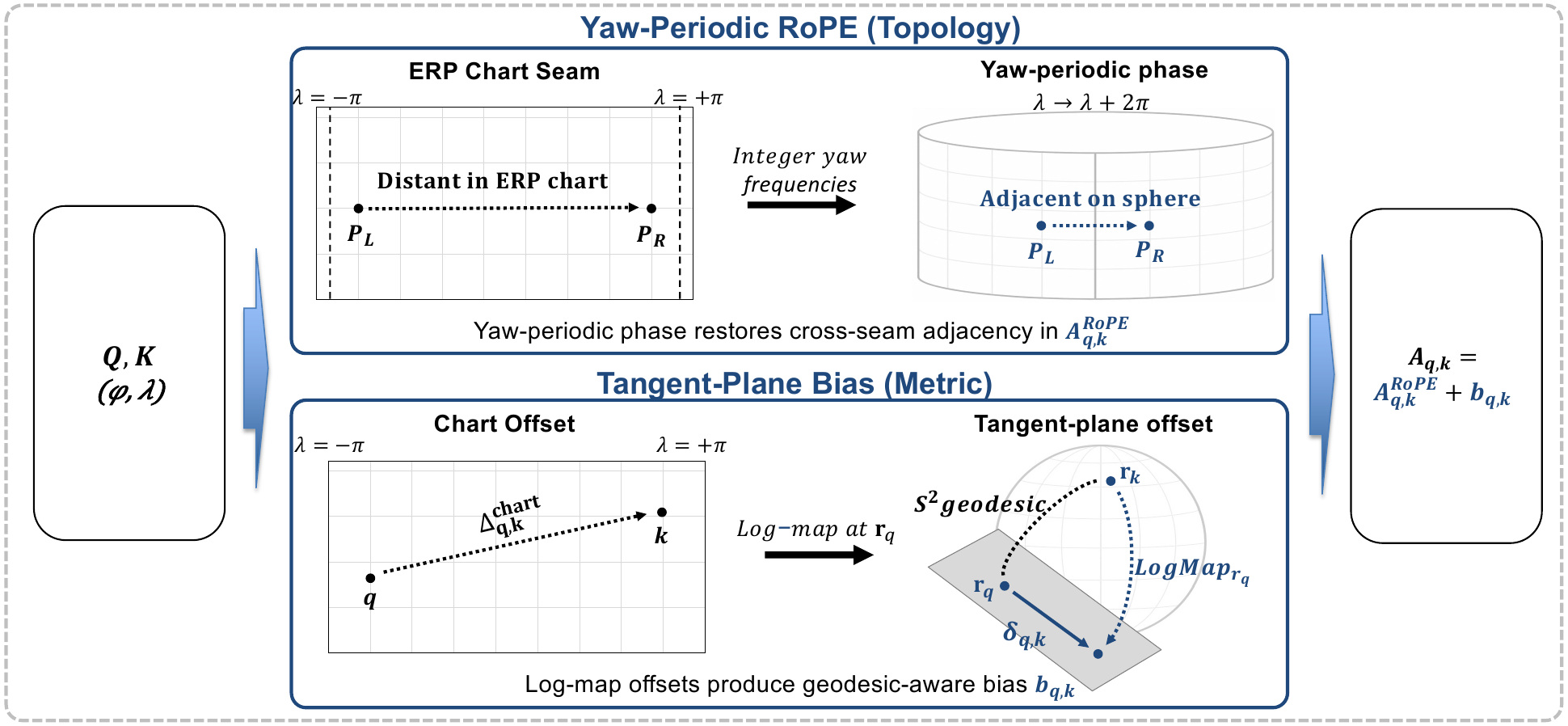}
\caption{\textbf{Spherical Positional Attention (SPA).}
From coarse query/key tokens $Q,K$ with sphere coordinates $(\varphi,\lambda)$,
SPA forms two sphere-aware positional terms that merge at the pre-softmax
attention logit. \textbf{(top)} Yaw-periodic RoPE makes the relative phase
strictly $2\pi$-periodic in longitude, so seam-neighbor tokens remain adjacent
across the ERP boundary. \textbf{(bottom)} Tangent-Plane Bias, instantiating the
CPB framework~\cite{swinv2} on the sphere, replaces the planar chart
offset $\Delta^{\mathrm{chart}}_{q,k}$ with the spherical log-map offset
$\boldsymbol{\delta}_{q,k}$ computed in the query tangent plane at $\mathbf{r}_q$,
producing a geodesic-aware additive bias $b_{q,k}$. The resulting logit is
$A_{q,k}=A^{\mathrm{RoPE}}_{q,k}+b_{q,k}$ (Sec.~\ref{sec:method-spa}).}
\label{fig:spa_module}
\end{figure}

\subsection{Spherical Positional Attention (SPA)}
\label{sec:method-spa}

SPA modifies the relative-position component of the coarse attention
layers via two parts (illustrated in Fig.~\ref{fig:spa_module}):
a yaw-periodic RoPE for longitudinal seam consistency, and a
tangent-plane bias for the local pairwise metric.

\noindent\textbf{(i) Yaw-Periodic RoPE.}
On ERP a yaw rotation of the camera (looking left/right about the vertical
axis) shifts every longitude $\lambda\!\to\!\lambda+\delta$ and wraps across the
$\pm\pi$ seam, so a seam-consistent positional encoding must be periodic in
$\lambda$.
Standard planar RoPE~\cite{roformer} uses geometric frequencies $\omega_i$ that are
not constrained to integer harmonics. Applying them to the ERP longitude
$\lambda \in [-\pi,\pi)$ would shift the relative phase
$\omega_i(\lambda_q-\lambda_k)$ by $2\pi\omega_i$---generally not a multiple
of $2\pi$---when one token's stored longitude wraps across the seam,
violating the chart's longitudinal periodicity.
Integer harmonics are in fact the only frequencies consistent with this
$2\pi$-periodicity (Supp.\ Sec.~H analyzes the resulting frequency resolution).

We split each attention head's $d_h$ channels into two halves of $d_h/2$ channels. 
The first half encodes latitude with the standard geometric schedule,
since latitude does not wrap; 
the second half encodes longitude.
As in standard RoPE, channels are rotated in pairs,
so the longitude half has $d_h/4$ pairs,
to which we assign integer frequencies $\omega_i^{\lambda} = i$.
For a query at latitude/longitude
$(\varphi_q,\lambda_q)\in[-\pi/2,\pi/2]\times[-\pi,\pi)$ and
a key at $(\varphi_k,\lambda_k)$,
the longitudinal relative phase of the $i$-th pair becomes
\begin{equation}
  \Delta\theta_i^{\lambda} = i\cdot(\lambda_q-\lambda_k),
  \qquad i = 1,\dots,d_h/4 .
  \label{eq:rel-phase}
\end{equation}
Because these longitudinal frequencies are integers,
a seam wrap ($\lambda\to\lambda+2\pi$) shifts each relative phase by a multiple of $2\pi$.
This leaves the rotation matrix $\mathbf{R}(\Delta\theta_i^{\lambda})$ of each channel pair unchanged,
making the RoPE-induced positional term strictly yaw-periodic.

\noindent\textbf{(ii) Tangent-Plane Bias (TPB).}
A relative-position bias is intended to calibrate spatial relations between
feature locations; on ERP the chart-pixel offset used by standard biases does
not correspond to uniform geodesic offsets across latitude, so a planar bias can
favor chart proximity over true spherical proximity. TPB corrects the metric of this bias. It
instantiates the Continuous Position Bias (CPB)
framework~\cite{swinv2} on the sphere. TPB maps a spherical offset to
2-D tangent-plane coordinates, Fourier-encodes them, and uses a shared MLP
to produce an additive attention bias. We adopt CPB's continuous-coordinate
MLP formulation; for SCCM, we use a scalar output initialized to zero and
replace planar pixel offsets with spherical log-map offsets.

Let $\mathbf{B}_{\mathbf{r}_q}\!\in\!\mathbb{R}^{3\times2}$ be an orthonormal
tangent basis at the query ray $\mathbf{r}_q$ (Eq.~\eqref{eq:ray}), aligned with
the ERP latitude/longitude directions; a near-pole-stable construction and
antipodal handling are deferred to Supp.\ Sec.~A. For every query we precompute
the 2-D geodesic offset of every key via the spherical log-map into this basis:
\begin{equation}
\boldsymbol{\delta}_{q,k}
= \mathbf{B}_{\mathbf{r}_q}^{\!\top}
\mathrm{LogMap}_{\mathbf{r}_q}(\mathbf{r}_k) \in \mathbb{R}^2 ,
\label{eq:tangent-delta}
\end{equation}
a spherical log-map offset that replaces the planar pixel offset. It is
encoded as sine/cosine Fourier features~\cite{tancik2020fourier} ($\gamma$)
and mapped by the shared MLP
to a scalar bias added to the pre-softmax logit $A^{\mathrm{RoPE}}_{q,k}$ (the
scaled RoPE-rotated query--key dot product):
\begin{equation}
b_{q,k}
= \mathrm{MLP}\!\bigl(\gamma(\boldsymbol{\delta}_{q,k})\bigr),
\qquad
A_{q,k} = A^{\mathrm{RoPE}}_{q,k} + b_{q,k}.
\label{eq:tangent-bias}
\end{equation}
Crucially, $\boldsymbol{\delta}_{q,k}$ depends only on the two grid locations'
chart coordinates---not on image content or pose---so it is a content-independent
spherical prior, precomputed once and shared across heads, SPA layers, and self-
and cross-attention; only the MLP and one latitude-scaling scalar are
learned (encoding details in Supp.\ Sec.~A).

\begin{figure}[!t]
\centering
\includegraphics[width=0.88\textwidth]{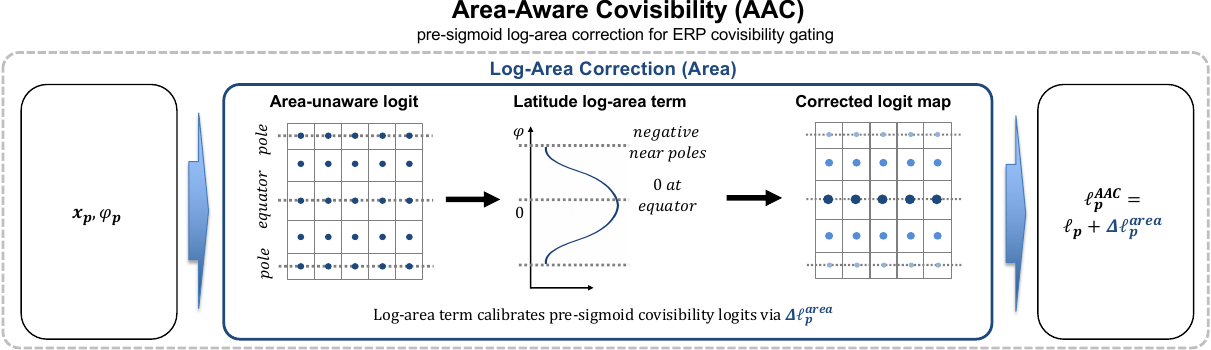}

\caption{\textbf{Area-Aware Covisibility (AAC).}
Given a per-pixel feature $\mathbf{x}_p$ and latitude $\varphi_p$, AAC adds a
latitude-dependent log-area term to the covisibility logit before the sigmoid:
$\ell^{\mathrm{AAC}}_p = \ell_p + \alpha_{\mathrm{LAC}}\log\max(\cos\varphi_p,\epsilon)$.
\textbf{(left)} the bare logit $\ell_p\!=\!\mathrm{MLP}(\mathbf{x}_p)$ is
area-unaware, whereas ERP pixels represent smaller spherical area near the
poles; \textbf{(middle)} the log-area term is $0$ at the equator and negative
toward the poles; \textbf{(right)} after sigmoid, the corrected logit yields an area-aware gate
that down-weights pixel-overrepresented polar candidates before dual-softmax matching, with learnable
$\alpha_{\mathrm{LAC}}\!=\!\mathrm{softplus}(\alpha_{\mathrm{raw}})$ (one per
image side, init\,$=\!1$). See Sec.~\ref{sec:method-aac}.}
\label{fig:aac_module}
\end{figure}

\subsection{Area-Aware Covisibility (AAC)}
\label{sec:method-aac}

AAC curbs the tendency of ERP's pixel-uniform sampling to over-represent
polar candidates at the covisibility gating stage. Concretely, AAC augments
the covisibility head with a single Log-Area Correction (LAC) term
defined below; the head structure itself is otherwise identical to
the baseline (chart-naïve) covisibility head.
The contribution of AAC is not learning a new area function---the ERP area
element is known analytically---but \emph{placing} this area prior before the
sigmoid, where the model decides whether a pixel should participate in matching.
Applying the same factor after dual-softmax, or only in the loss, would not
directly calibrate the pre-sigmoid covisibility decision.

\noindent\textbf{Log-Area Correction (LAC).}
At the covisibility gating stage, the covisibility head produces a per-pixel
covisibility logit $\ell = \mathrm{MLP}(\mathbf{x})$, where $\mathbf{x}$ is a
learned feature vector. This logit is mapped by a sigmoid to a soft gating
probability $\mu=\sigma(\ell)$ used to filter spurious matches before the
dual-softmax step. This gate is passed to the refiner as the coarse certainty
signal, so area-correcting $\mu$ affects the certainty refined downstream.

To correct this polar over-representation, we augment the bare covisibility logit for each pixel
feature $\mathbf{x}_p \in \mathbb{R}^d$ at latitude $\varphi_p$ with a
learnable LAC term (Fig.~\ref{fig:aac_module}). The corrected gate is
$\mu_p = \sigma(\ell^{\mathrm{AAC}}_p)$, where
\begin{equation}
\ell^{\mathrm{AAC}}_p
= \mathrm{MLP}(\mathbf{x}_p) + \alpha_{\mathrm{LAC}} \cdot
\log \max\!\bigl(\cos\varphi_p,\, \epsilon\bigr),
\qquad
\epsilon = 10^{-3}.
\label{eq:lac}
\end{equation}
The logarithm converts the multiplicative area factor into an additive logit
correction, matching the pre-sigmoid interface of the covisibility gate (we
abbreviate the two terms as $\ell_p + \Delta\ell^{\mathrm{area}}_p$ in
Fig.~\ref{fig:aac_module}).
The clamp $\epsilon$ is used only for numerical stability near the poles;
sensitivity is reported in Supp. Sec.~K.
In contrast to input- or refinement-level spherical embeddings, LAC enters the \emph{pre-sigmoid} covisibility logit
as a learnable scalar
$\alpha_{\mathrm{LAC}}\!=\!\mathrm{softplus}(\alpha_{\mathrm{raw}})\!>\!0$ on the
area term (one per image side $v\!\in\!\{A,B\}$), initialized at
$\alpha_{\mathrm{LAC}}\!=\!1$ so that the correction starts exactly at the
analytic ERP-to-sphere area Jacobian $|J|\!=\!\cos\varphi$.
The learned strengths remain near this analytic value (Supp.\ Sec.~K).

\begin{table}[!t]
\centering
\caption{%
\textbf{Main full-system comparison} on (a) Matterport3D, (b) zero-shot
Stanford2D3D, and (c) outdoor Holo360D, where all three matchers are trained on Holo360D
from their Matterport3D checkpoints with a shared training budget (Supp.~Sec.~P). All methods are evaluated
end-to-end under the same ERP dense matching metrics. Retrained/controlled rows in (a,\,b) share the protocol of
Sec.~\ref{sec:exp-setup}, while external baselines use their released public
weights. The mechanism-isolated contribution of the sphere-aware priors,
measured as chart-naïve\,$\to$\,SCCM under a fixed scaffold, is reported in
Tab.~\ref{tab:body_ablation}. Stanford2D3D uses the overlap range
$0.30\!\le\!\mathrm{ov}\!\le\!0.80$; details in Supp.\ Sec.~B.
Best \textbf{bold}, second \underline{underlined}.
$^\dagger$SphereGlue~\cite{sphereglue} is sparse and not directly comparable to
dense methods.}
\label{tab:main}
\setlength{\tabcolsep}{3pt}
\renewcommand{\arraystretch}{1.1}
\footnotesize
\resizebox{\columnwidth}{!}{%
\begin{tabular}{l l c c c c c}
\toprule
\textbf{Method} & \textbf{Train data}
 & \textbf{PCK@1}$^\circ\!\uparrow$
 & \textbf{PCK@3}$^\circ\!\uparrow$
 & \textbf{PCK@5}$^\circ\!\uparrow$
 & \textbf{MAE}$^\circ\!\downarrow$
 & \textbf{Med.}$^\circ\!\downarrow$ \\
\midrule
\rowcolor{gray!18}\multicolumn{7}{l}{\textbf{(a) Matterport3D test --- in-distribution ($15{,}682$ pairs)}} \\
\multicolumn{7}{l}{Perspective baselines (trained on perspective, zero-shot on ERP)} \\
\quad RoMa~V1~\cite{roma}              & ScanNet~\cite{scannet} (indoor, persp.) & 0.023 & 0.058 & 0.081 & 57.46 & 49.74 \\
\quad RoMa~V2~\cite{romav2}            & Mixed (incl. indoor, persp.) & 0.040 & 0.126 & 0.201 & 36.10 & 18.09 \\
\multicolumn{7}{l}{ERP-native baselines (released checkpoints, not retrained)} \\
\quad SphereGlue$^\dagger$~\cite{sphereglue}     & Custom Synth.\ (indoor, ERP) & 0.028 & 0.072 & 0.095 & 64.50 & 65.45 \\
\quad EDM~\cite{edm}                   & MP3D (indoor, ERP)        & 0.163 & 0.400 & 0.509 & 16.78 & 4.79 \\
\multicolumn{7}{l}{ERP-retrained baseline} \\
\quad RoMa~V1 (ERP-retrained)~\cite{roma}        & MP3D (indoor, ERP)        & 0.198 & 0.538 & 0.711 & 6.60 & 2.76 \\
\multicolumn{7}{l}{Fixed-scaffold control (ours, no sphere geometry)} \\
\quad chart-naïve (cross-attn$+$DS, no PE) & MP3D (indoor, ERP) & \underline{0.230} & \underline{0.609} & \underline{0.769} & \underline{5.68} & \underline{2.23} \\
\multicolumn{7}{l}{Ours} \\
\quad\textbf{SCCM} (full) & MP3D (indoor, ERP) & \textbf{0.275} & \textbf{0.665} & \textbf{0.806} & \textbf{5.36} & \textbf{1.88} \\
\midrule
\rowcolor{gray!18}\multicolumn{7}{l}{\textbf{(b) Stanford2D3D test --- out-of-distribution / zero-shot ($8{,}744$ pairs)}} \\
\multicolumn{7}{l}{Perspective baselines (trained on perspective, zero-shot on ERP)} \\
\quad RoMa~V1~\cite{roma}              & ScanNet (indoor, persp.) & 0.022 & 0.051 & 0.072 & 64.08 & 58.99 \\
\quad RoMa~V2~\cite{romav2}            & Mixed (incl. indoor, persp.) & 0.043 & 0.125 & 0.201 & 42.15 & 18.59 \\
\multicolumn{7}{l}{ERP-native baselines (released checkpoints, not retrained)} \\
\quad SphereGlue$^\dagger$~\cite{sphereglue}     & Custom Synth.\ (indoor, ERP) & 0.034 & 0.072 & 0.093 & 67.96 & 62.78 \\
\quad EDM~\cite{edm}                   & MP3D (indoor, ERP)      & 0.104 & 0.262 & 0.355 & 35.70 & 10.69 \\
\multicolumn{7}{l}{ERP-retrained baseline} \\
\quad RoMa~V1 (ERP-retrained)~\cite{roma}        & MP3D (indoor, ERP)      & 0.167 & 0.452 & \underline{0.628} & \underline{17.80} & 3.45 \\
\multicolumn{7}{l}{Fixed-scaffold control (ours, no sphere geometry)} \\
\quad chart-naïve (cross-attn$+$DS, no PE) & MP3D (indoor, ERP) & \underline{0.179} & \underline{0.475} & 0.627 & 18.86 & \underline{3.26} \\
\multicolumn{7}{l}{Ours} \\
\quad\textbf{SCCM} (full) & MP3D (indoor, ERP) & \textbf{0.229} & \textbf{0.564} & \textbf{0.697} & \textbf{17.09} & \textbf{2.45} \\
\midrule
\rowcolor{gray!18}\multicolumn{7}{l}{\textbf{(c) Holo360D test --- outdoor, in-the-wild ($8{,}000$ pairs)}} \\
\multicolumn{7}{l}{Shared training budget (ours; Supp.~Sec.~P)} \\
\quad RoMa~V1 (ERP-retrained)~\cite{roma} & Holo360D (outdoor) & 0.322 & 0.592 & 0.726 & 5.66 & 2.12 \\
\quad chart-naïve (cross-attn$+$DS, no PE) & Holo360D (outdoor) & \underline{0.331} & \underline{0.626} & \underline{0.761} & \underline{5.02} & \underline{1.93} \\
\quad \textbf{SCCM} (full) & Holo360D (outdoor) & \textbf{0.357} & \textbf{0.645} & \textbf{0.776} & \textbf{4.94} & \textbf{1.76} \\
\bottomrule
\end{tabular}}
\end{table}

\section{Experiments}
\label{sec:experiments}

\subsection{Experimental Setup}
\label{sec:exp-setup}

\noindent\textbf{Implementation.}
All models retrained on Matterport3D share a fixed $1$\,M-sample protocol
($448\!\times\!896$ ERP inputs, frozen DINOv2-Large, original RoMa~V1 loss),
data split, and evaluator. Each main ablation configuration is trained once
with a common initialization seed. Three runs per endpoint assess training
variability (s.d.\ $\le\!0.6$\,pp; Supp.~Sec.~R).
The $+0.9$\,pp LAC step is a single-run estimate.

\noindent\textbf{Datasets.}
We train on Matterport3D~\cite{matterport3d} (MP3D, indoor) using its official
benchmark split, evaluate zero-shot on Stanford2D3D~\cite{stanford2d3d}
(indoor), and continue training the Matterport3D models on Holo360D~\cite{holo360d} (outdoor; Sec.~\ref{sec:exp-comparison}). All retrained models
use $H\!\times\!W\!=\!448\!\times\!896$ ERP
inputs (matching the DINOv2 patch grid), with native panoramas resized to this
resolution (native dataset resolutions in Supp.\ Sec.~A/B).
Released checkpoints are each evaluated at their own native resolution
($320\!\times\!640$ for EDM), with ground truth and metrics on that grid (resolution-matched control for EDM
in Supp.~Sec.~R).
SCCM assumes gravity-aligned (upright) ERP, as in the Matterport3D and
Stanford2D3D benchmarks; robustness to camera tilt is analyzed in Supp.~Sec.~F.

\noindent\textbf{Evaluation metrics.}
We evaluate angular error on the sphere,
$\theta = \arccos(\mathbf{r}_{\mathrm{pred}} \!\cdot\! \mathbf{r}_{\mathrm{gt}})$,
rather than pixel-space endpoint error (biased by ERP's $\cos\varphi$
anisotropy), and report PCK (the fraction of valid pixels below angular
thresholds) at $\{1,3,5\}^\circ$ with mean (MAE) and median angular error;
PCK@$1^\circ$ is the primary strict-precision metric.

Full training protocol, dataset construction, and complexity details are in
Supp. Secs.~A, B, and~L.

\subsection{Benchmark Comparison}
\label{sec:exp-comparison}

\noindent\textbf{Model groups.}
Under the shared evaluation protocol of Sec.~\ref{sec:exp-setup} we compare: released
\emph{perspective baselines} (RoMa~V1~\cite{roma}, RoMa~V2~\cite{romav2},
zero-shot on ERP), \emph{ERP-native baselines} (sparse
SphereGlue~\cite{sphereglue}, dense EDM~\cite{edm}), an \emph{ERP-retrained
RoMa~V1} (same protocol and budget as ours), our \emph{chart-naïve} scaffold (cross-attention $+$ dual-softmax, no
positional encoding), and \emph{SCCM} (chart-naïve $+$ SPA $+$ AAC). RoMa~V1
thus plays three roles: zero-shot checkpoint, ERP-retrained baseline, and the
base architecture for our controlled study. The ablation
(Sec.~\ref{sec:exp-ablation}) adds two mechanism-only positional-encoding (PE)
controls on that scaffold---standard geometric-frequency RoPE and EDM's
input-side absolute PE---re-implemented as the mechanism only, not the full EDM
system.
EDM is evaluated from its released, Matterport3D-trained checkpoint---its training code is
not public (Supp.~Sec.~G)---so that comparison is matched at evaluation but not
at training recipe. Four further matchers (DKM, LoFTR, MASt3R, VGGT~\cite{vggt}) evaluated
zero-shot on ERP all fall below EDM (Supp.~Sec.~Q).

\begin{figure}[!tbp]
\centering
\includegraphics[width=\textwidth]{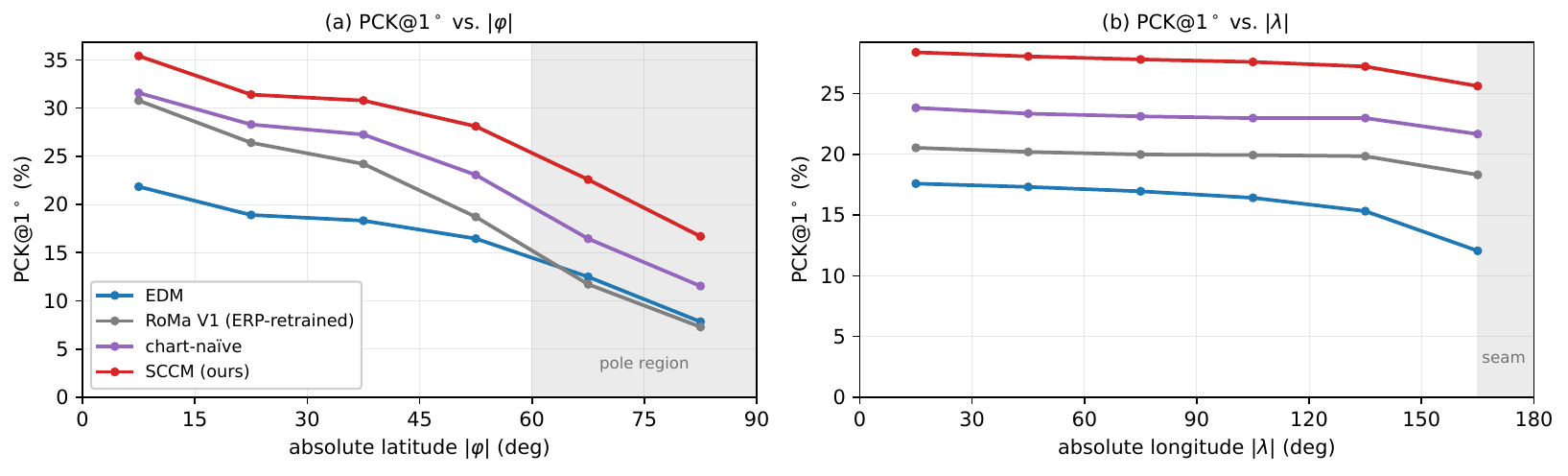}
\caption{\textbf{Precision analysis on Matterport3D test}
(full test set, $15{,}682$ pairs; common $448\!\times\!896$ evaluation grid).
(a)~PCK@$1^\circ$ by absolute latitude $|\varphi|$: SCCM is consistently
highest across latitude bands, with a larger relative margin toward the poles (shaded).
(b)~PCK@$1^\circ$ by absolute longitude $|\lambda|$: SCCM shows a smaller drop
near the ERP seam ($|\lambda|\!=\!180^\circ$, shaded) than EDM, while the
chart-naïve scaffold lies between EDM and SCCM.}
\label{fig:pck_vs_lat}
\end{figure}

\noindent\textbf{Decomposing the gain.}
We report two distinct gains on Matterport3D (Tab.~\ref{tab:main}a): a
\emph{scaffold gain} from replacing RoMa~V1's GP coarse stage with our
chart-naïve R1 ($0.198\!\to\!0.230$ PCK@$1^\circ$, $+3.2$\,pp), and a
\emph{spherical-prior gain} from adding the three sphere-derived modules on the
fixed R1 scaffold ($0.230\!\to\!0.275$, $+4.5$\,pp; Sec.~\ref{sec:exp-ablation}).
The latter is the mechanism-isolated contribution of SCCM and the central claim
of this paper; the full-system numbers below combine both.

\begin{figure}[!tbp]
\centering
\includegraphics[width=\textwidth]{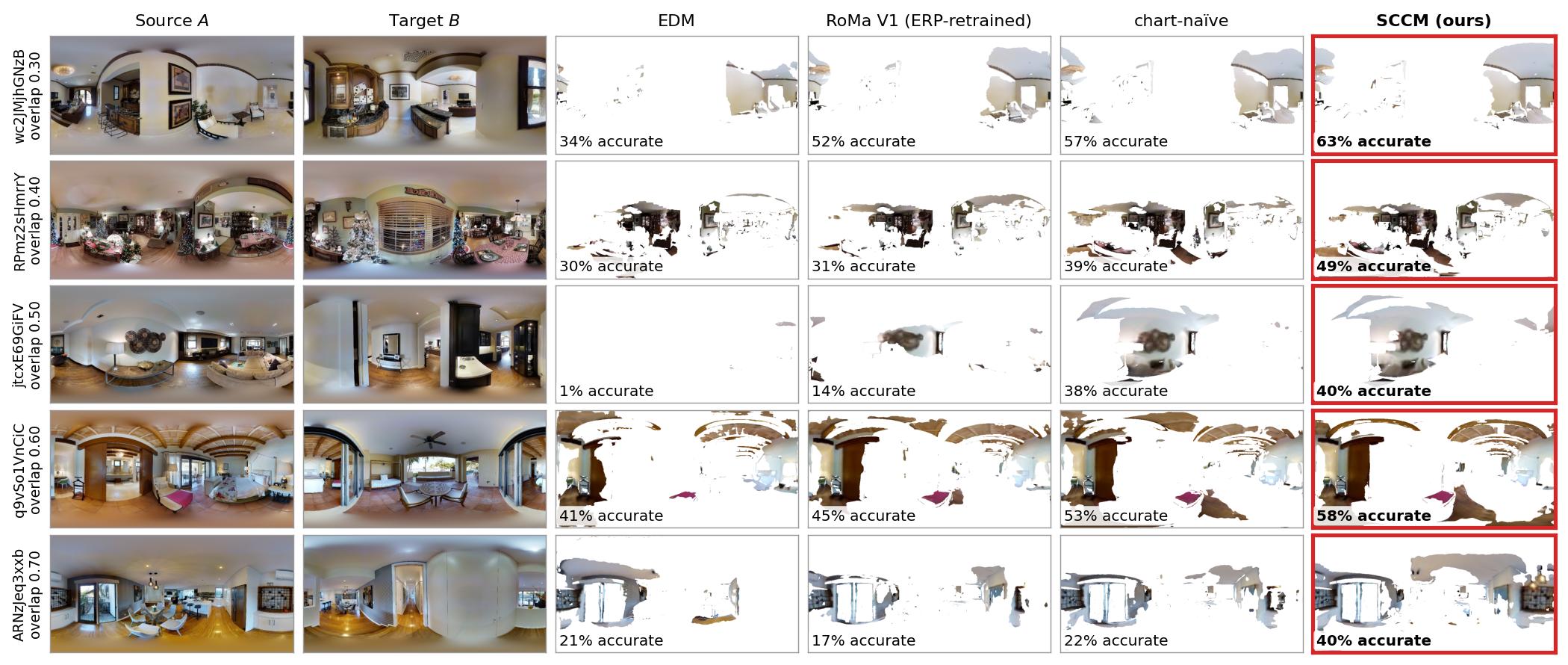}
\caption{\textbf{Qualitative comparison on Matterport3D test.}
Each method panel shows the predicted warp masked by correctness: white denotes
inaccurate or non-covisible pixels. The reported percentage is the fraction of
GT-covisible pixels with angular error below $1^\circ$. SCCM recovers larger
accurate regions than the dense baselines (EDM, RoMa~V1 (ERP-retrained)) and the
chart-naïve scaffold across representative overlap levels, including
high-latitude ceiling/floor regions.}
\label{fig:qual}
\end{figure}

\noindent\textbf{In-distribution: Matterport3D test.}
Tab.~\ref{tab:main}(a) reports results on the official Matterport3D test split.
Zero-shot perspective baselines (RoMa~V1, RoMa~V2) stay below $0.05$
PCK@$1^\circ$, reflecting ERP distortion compounded by the perspective-to-ERP
domain shift. Among prior published ERP methods, the ERP-native EDM is strongest
(PCK@$1^\circ\!=\!0.163$, MAE$\,=\,16.78^\circ$; the sparse SphereGlue is not
directly comparable), and our ERP-retrained RoMa~V1 ($0.198$) already
surpasses it. Our chart-na\"ive backbone surpasses EDM \emph{without any
sphere-aware encoding} ($0.230$ vs.\ $0.163$), and positional-encoding controls
on it change little (Tab.~\ref{tab:body_ablation}): it is a
strong matching substrate, and generic positional encodings alone do not
explain the SCCM gain. SCCM reaches PCK@$1^\circ\!=\!0.275$, $+11.2$\,pp over
EDM and $+4.5$\,pp over the chart-na\"ive backbone (Tab.~\ref{tab:main}a).
Fig.~\ref{fig:qual} shows the gap qualitatively across overlap levels.

\begin{table}[!b]
\centering
\caption{\textbf{Controlled ablation on the fixed chart-naïve scaffold}
(Matterport3D test). R2a--R3 cumulatively add yaw-periodic RoPE, TPB, and
LAC/AAC to the no-prior scaffold R1; the two $^{\S}$ rows are positional-encoding
controls on R1. Values in parentheses on PCK@$1^\circ$ and MAE denote the
marginal change from the previous chain row (PCK in pp, MAE in degrees), revealing the
modules' functional roles: yaw-periodic RoPE drives strict precision, TPB
reduces the angular-error tail, and LAC adds area-gated precision. Chain
marginals on PCK@$1^\circ$ fold the TPB$\times$LAC interaction into the final
row (factorial decomposition in Supp.\ Sec.~M). Protocol as in
Sec.~\ref{sec:exp-setup}. Best in \textbf{bold}.}
\label{tab:body_ablation}
\setlength{\tabcolsep}{4pt}
\renewcommand{\arraystretch}{1.1}
\footnotesize
\resizebox{\columnwidth}{!}{%
\begin{tabular}{l ccc cc l}
\toprule
 & \multicolumn{3}{c}{\textbf{PCK}$^\circ$ ($\uparrow$)}
 & \multicolumn{2}{c}{\textbf{Error}$^\circ$ ($\downarrow$)} & \\
\cmidrule(lr){2-4}\cmidrule(lr){5-6}
\textbf{Row} & \textbf{@1} & \textbf{@3} & \textbf{@5} & \textbf{MAE} & \textbf{Med.} & \textbf{Primary role} \\
\midrule
(R1) chart-naïve (no PE)                    & 0.230 & 0.609 & 0.769 & 5.68 & 2.23 & planar scaffold \\
\quad\textit{ctrl: EDM abs.\ PE}$^{\S}$     & 0.224 & 0.593 & 0.756 & 5.97 & 2.31 & \textit{(input-PE control)} \\
\quad\textit{ctrl: standard RoPE}$^{\S}$    & 0.236 & 0.613 & 0.764 & 6.79 & 2.17 & \textit{(rel-PE control)} \\
(R2a) $+$ yaw-periodic RoPE   & $0.267$\,($+3.7$) & 0.653 & 0.796 & $5.87$\,($+0.19$) & 1.93 & strict precision \\
(R2b) $+$ TPB (full SPA)      & $0.266$\,($-0.1$) & 0.652 & 0.799 & $5.58$\,($-0.29$) & 1.95 & error-tail (MAE) \\
\textbf{(R3) $+$ LAC $\,\equiv\,$ SCCM} & $\mathbf{0.275}$\,($+0.9$) & \textbf{0.665} & \textbf{0.806} & $\mathbf{5.36}$\,($-0.22$) & \textbf{1.88} & area-gated precision \\
\bottomrule
\end{tabular}}
\end{table}

\noindent\textbf{Absolute vs.\ pairwise positional encoding.}
The EDM absolute PE control re-implements EDM's input-side absolute
sphere-coordinate embedding on the chart-na\"ive backbone, testing whether
providing sphere coordinates to a strong planar backbone suffices. It performs
slightly below the backbone alone (Tab.~\ref{tab:body_ablation}), so SCCM's lead is essentially
unchanged against it. Absolute encoding assigns coordinates to each location
independently, without explicitly imposing SPA's relative query--key geometry.
The largest ablation step occurs at R1$\to$R2a ($+3.7$\,pp). This control tests the input-side absolute-PE
mechanism under the same scaffold, not EDM as a full system.

\noindent\textbf{Out-of-distribution: Stanford2D3D.}
Tab.~\ref{tab:main}(b) reports zero-shot evaluation on the filtered
Stanford2D3D test-pair set ($8{,}744$ pairs), using overlap
$0.30\!\le\!\mathrm{ov}\!\le\!0.80$ for all methods to include low-covisibility
pairs and exclude near-duplicates (Supp.\ Sec.~B). SCCM exceeds every baseline
(Tab.~\ref{tab:main}b), so the sphere-aware
advantage transfers across panoramic datasets.

\noindent\textbf{Outdoor.} For Tab.~\ref{tab:main}(c), RoMa~V1
(ERP-retrained), chart-na\"ive, and SCCM are trained from their MP3D checkpoints
with a shared training budget on Holo360D~\cite{holo360d}, an in-the-wild handheld
LiDAR$+$360$^\circ$ dataset ($2.2\times$ deeper than MP3D, median relative tilt
$12^\circ$; scene-disjoint 5/3/4 splits, $8{,}000$ test pairs). SCCM leads on
every metric, by $+2.6$/$+3.5$\,pp PCK@$1^\circ$ over chart-na\"ive/RoMa~V1;
zero-shot outdoor transfer is analyzed in Supp.~Sec.~P.

\noindent\textbf{Latitude-wise behavior.}
Fig.~\ref{fig:pck_vs_lat}(a) reports PCK@$1^\circ$ by absolute latitude on
Matterport3D test. Baselines degrade steeply as $|\varphi|$ grows, since the
same chart offset spans different geodesic distances across latitudes. SCCM
improves PCK@$1^\circ$ at every latitude, with the margin growing toward the
poles ($\approx\!1.45\times$ the chart-na\"ive scaffold at the highest band vs.\
$\approx\!1.12\times$ at the equator), and remains more robust than EDM across the
longitudinal seam (Fig.~\ref{fig:pck_vs_lat}b). Because pixel-uniform
evaluation over-represents polar pixels, we also weight pixels by $\cos\varphi$;
the chart-na\"ive\,$\to$\,SCCM improvement persists on every metric
(PCK@$1^\circ$ $+3.9$\,pp area-weighted vs.\ $+4.5$\,pp; Supp.\ Sec.~E).

\subsection{Ablation Study}
\label{sec:exp-ablation}

Tab.~\ref{tab:body_ablation} isolates the sphere-aware modules on the
fixed chart-na\"ive scaffold R1 (cross-attention $+$ dual-softmax), whose
standing against all baselines is established in Sec.~\ref{sec:exp-comparison}.
R2a--R3 progressively add yaw-periodic RoPE, TPB, and AAC with all other
settings held constant, so the $+4.5$\,pp PCK@$1^\circ$ gain from R1 to R3
isolates the spherical modules from the scaffold replacement (full MP3D test
set, $15{,}682$ pairs, ${\sim}3.0\mathrm{B}$ pixels per checkpoint).

\noindent\textbf{Module contributions.}
The three modules act along distinct axes
(Tab.~\ref{tab:body_ablation} lists marginal changes and primary roles;
further analysis in Supp.\ Sec.~M). \emph{Yaw-periodic RoPE} supplies most of the precision
gain---$+3.7$\,pp PCK@$1^\circ$ over R1---and the gain is specific to the $2\pi$-periodic design:
substituting standard geometric-frequency RoPE
(Tab.~\ref{tab:body_ablation},~$^{\S}$) yields only $+0.6$\,pp and
\emph{degrades} MAE ($5.68^\circ\!\to\!6.79^\circ$), supporting longitudinal
periodicity as the key factor rather than generic relative encoding. \emph{TPB} leaves strict PCK essentially unchanged ($-0.1$\,pp) while reducing
the angular-error tail (MAE $5.87^\circ\!\to\!5.58^\circ$; tail quantiles in
Supp.\ Sec.~M); its contribution is
metric calibration---conditioning the pre-softmax bias on the true geodesic
offset---complementary to the precision gain of yaw-periodic RoPE.
\emph{LAC (AAC)} adds the log-area term
$\alpha_{\mathrm{LAC}}\log\max(\cos\varphi,\epsilon)$
to the pre-sigmoid covisibility logit, calibrating pixel-overrepresented polar
candidates through the known area prior: it lowers MAE ($-0.22^\circ$) and
median ($-0.07^\circ$) and lifts PCK@$1^\circ$ by $+0.9$\,pp
(Tab.~\ref{tab:body_ablation}). A $2\times2$ factorial with a RoPE$+$LAC
control (Supp.\ Sec.~M) attributes this strict-precision step to the
TPB$\times$LAC \emph{interaction} rather than to LAC alone: neither prior
improves PCK@$1^\circ$ in isolation, while both reduce MAE.
Their combination improves strict precision, consistent with the complementary
roles of the area-corrected gate and the geodesic pairwise bias. The learned $\alpha_{\mathrm{LAC}}$ settles near
the analytic $\cos\varphi$ Jacobian; setting $\alpha\!=\!1$ at inference
changes the reported metrics by at most $10^{-4}$ (Supp.\ Sec.~K).

\noindent\textbf{Coarse-stage leverage.}
Coarse-anchor quality is thus the key leverage point under this
scaffold: the full $+4.5$\,pp PCK@$1^\circ$ gain is obtained by changing only
the coarse-stage modules, with the encoder, refiner architecture, loss, and
training protocol fixed. A refinement-stage diagnostic on the full test split
(Supp.~Sec.~O) supports this: SCCM's advantage is already present at the
coarse-anchor stage ($+4.1$\,pp) and remains stable through refinement
($+4.5$\,pp at the final output). The shared refiner architecture yields a similar
coarse-to-final lift for R1 and SCCM ($+10.8$ vs.\ $+11.2$\,pp),
supporting coarse-anchor formation as the principal source of the gain.

\subsection{Downstream Tasks}
\label{sec:exp-downstream}

We evaluate whether the dense matching gains translate to downstream geometry on
Matterport3D under a single unified evaluator (Tab.~\ref{tab:downstream_main}).
Both tasks use each matcher's per-pixel certainty: pose samples ray
matches by certainty, and reconstruction applies one fixed threshold
($\tau\!=\!0.5$) to every method.
For \textbf{relative pose} (essential-matrix $+$ RANSAC on certainty-weighted
ray matches), SCCM improves Pose AUC at every threshold over the chart-naïve scaffold (Tab.~\ref{tab:downstream_main}).
An estimator swap to 360-8PA~\cite{solarte360} with fixed matches and
inliers changes AUC@$5^\circ$ by $\le\!0.01$\,pp for the four methods
in Supp.~Sec.~R. For
\textbf{3D reconstruction}, we triangulate under the GT pose to isolate
correspondence quality from pose-estimation errors; SCCM leads on \emph{all}
seven accuracy, completeness, and F-score metrics, ahead of the
chart-naïve scaffold, RoMa~V1 (ERP-retrained), and EDM. Ungated
reconstruction (Supp.\ Sec.~N) shows the same trend except for mean accuracy;
details and top-down visualizations in Supp.\ Sec.~C.

\begin{table*}[t]
\centering
\caption{\textbf{Downstream geometry on Matterport3D under a unified evaluator.}
We report relative-pose AUC and certainty-gated reconstruction metrics.
Reconstruction triangulates matches under the GT pose with a fixed certainty
gate ($\tau\!=\!0.5$) for all methods; both tasks are evaluated on the $11{,}574$ test pairs where every method yields
$\ge\!200$ confident matches; details in Supp.\ Sec.~C. Best
\textbf{bold}, second \underline{underlined}.}
\label{tab:downstream_main}
\setlength{\tabcolsep}{2.7pt}
\scriptsize
\begin{tabular}{l ccc cc cc ccc}
\toprule
 & \multicolumn{3}{c}{\textbf{Pose AUC}$\,\uparrow$}
 & \multicolumn{2}{c}{\textbf{Acc} (m)$\,\downarrow$}
 & \multicolumn{2}{c}{\textbf{Comp} (m)$\,\downarrow$}
 & \multicolumn{3}{c}{\textbf{Recon.\ F}$\,\uparrow$} \\
\cmidrule(lr){2-4}\cmidrule(lr){5-6}\cmidrule(lr){7-8}\cmidrule(lr){9-11}
\textbf{Method} & $5^\circ$ & $10^\circ$ & $20^\circ$
 & Mean & Med. & Mean & Med. & @5 & @10 & @20 \\
\midrule
EDM~\cite{edm}                       & 7.97 & 17.86 & 29.51 & 1.452 & 0.470 & 0.685 & 0.336 & 14.5 & 27.5 & 42.7 \\
RoMa~V1 (retr.)~\cite{roma}          & 11.12 & 27.26 & 47.42 & 0.884 & 0.294 & 0.573 & 0.250 & 16.0 & 30.7 & 48.2 \\
chart-naïve                          & \underline{13.66} & \underline{31.50} & \underline{51.76}
                                     & \underline{0.787} & \underline{0.267} & \underline{0.512} & \underline{0.236}
                                     & \underline{17.4} & \underline{32.7} & \underline{50.6} \\
\textbf{SCCM} (ours)                 & \textbf{17.41} & \textbf{35.84} & \textbf{55.09}
                                     & \textbf{0.772} & \textbf{0.247} & \textbf{0.496} & \textbf{0.218}
                                     & \textbf{19.3} & \textbf{35.5} & \textbf{53.2} \\
\bottomrule
\end{tabular}
\end{table*}

\section{Conclusion}
\label{sec:conclusion}

We introduced SCCM, a coarse matcher that corrects the three geometric
distortions of equirectangular projection---topology, pairwise metric, and
per-pixel area---in the coarse-stage operations where they arise (attention
and covisibility gating), through lightweight modules that preserve the
encoder, refiner architecture, and loss. Under our unified ERP dense matching protocol, SCCM
outperforms all tested perspective-trained, ERP-retrained, and ERP-native
baselines on Matterport3D, transfers zero-shot to Stanford2D3D, and leads
on outdoor Holo360D when trained on it.
The priors can be adapted to other matchers with the corresponding attention
and covisibility-logit interfaces; GP-based coarse stages require attention
to be added before SPA can apply (Supp.~Sec.~G).

\noindent\textbf{Limitations and future work.}
SCCM assumes gravity-aligned ERP and does not explicitly handle camera tilt:
under synthetic camera-pitch perturbations all tested ERP matchers degrade
sharply (SCCM PCK@$1^\circ$: $0.273\!\to\!0.142$ at $10^\circ$ pitch,
$0.034$ at $30^\circ$, though still best at every angle; Supp.~Tab.~5), and
SCCM is not tilt-equivariant (failure case: Supp.~Sec.~N). It also inherits
the frozen encoder's (DINOv2-Large) limitations in texture-less,
photometrically ambiguous, or low-overlap regions. SCCM has been trained only on
Matterport3D (indoor) and Holo360D (outdoor); applied zero-shot to a further
outdoor corpus, the Holo360D-trained model leads only at $\le\!1^\circ$ (Supp.~Sec.~P). SCCM leaves the refiner sphere-na\"ive (Supp.~Sec.~O). Rotation-equivariant coarse
matching on $SO(3)$, ambiguity-aware encoders, and sphere-aware fine
refinement are future work.

\subsubsection*{\ackname}
This work was supported by the Institute of Information \& Communications
Technology Planning \& Evaluation (IITP) grant funded by the Korea government
(MSIT) (No.~RS-2023-00229833, Development of Intelligent Teleoperation
Technology for Cloud-Based Autonomous Vehicle Errors and Limit Situations).

%
%
\bibliographystyle{splncs04}
\bibliography{main}
\end{document}


\maketitle

\appendix

\section{Implementation Details}
\label{sec:supp-impl}

\noindent\textbf{Reproducibility.}
Tab.~\ref{tab:supp-protocol} summarizes the shared training protocol used for
all models retrained on Matterport3D~\cite{matterport3d} (the Holo360D training protocol is in
Sec.~\ref{sec:supp-holo360d}); SCCM changes only the coarse-matcher
modules, leaving the loss, encoder, and refiner architecture unchanged.

\begin{table}[ht]
\centering
\caption{\textbf{Shared training protocol} for all models retrained on
Matterport3D, used for the fixed-scaffold ablations.}
\label{tab:supp-protocol}
\footnotesize
\setlength{\tabcolsep}{6pt}
\renewcommand{\arraystretch}{1.2}
\begin{tabular}{@{}lp{0.66\textwidth}@{}}
\toprule
\textbf{Setting} & \textbf{Value} \\
\midrule
Optimizer & AdamW ($\beta_1{=}0.9$, $\beta_2{=}0.999$, wd $0.01$); non-zero LR only on trainable modules \\
LR schedule & base $1\!\times\!10^{-4}$, constant, $\times0.1$ at step $112{,}500$ ($90\%$ of the 125K steps) \\
Budget & $1$\,M samples ($\approx$125K steps); all rows identical \\
Loss & RoMa~\cite{roma} dense regression $+$ certainty BCE $+$ coarse-grid CE, unmodified (SCCM adds none) \\
GT corresp. & spherical warp from per-pixel depth $+$ relative pose (intrinsics-free ERP) \\
MP3D split & official $61/11/18$ scenes $=54{,}015/4{,}625/15{,}682$ pairs; native $1024\!\times\!512$ \\
Augmentation & pair-shared horizontal flip ($p=0.5$) and independent per-frame yaw $\theta_A,\theta_B\!\in\![0,2\pi)$ (EDM~\cite{edm} convention); no color/crop \\
Encoder & DINOv2-Large (patch 14), frozen; $448\!\times\!896\!\to\!32\!\times\!64$ tokens ($N{=}2048$) \\
Resolution & inputs resized to $448\!\times\!896$; metrics on the dense $448\!\times\!896$ refiner warp \\
Hardware & $8\!\times$RTX 4090 (24\,GB), batch $1$/GPU, bf16 forward; $\approx$22\,h/run \\
\bottomrule
\end{tabular}
\end{table}

\noindent\textbf{Numerical TPB details.}
For a ray $\mathbf{r}=(\cos\varphi\sin\lambda,\,\sin\varphi,\,\cos\varphi\cos\lambda)$,
the orthonormal tangent frame $\mathbf{B}_{\mathbf{r}}=[\mathbf{e}_\varphi\;\mathbf{e}_\lambda]$
(Sec.~4.1, main) is
\begin{equation}
\mathbf{e}_\varphi=\frac{\partial\mathbf{r}}{\partial\varphi}
=(-\sin\varphi\sin\lambda,\ \cos\varphi,\ -\sin\varphi\cos\lambda),
\qquad
\mathbf{e}_\lambda=\mathbf{e}_\varphi\times\mathbf{r} .
\end{equation}
$\mathbf{e}_\varphi$ has unit norm for every $(\varphi,\lambda)$, including the poles;
$\mathbf{e}_\lambda$ is automatically unit and orthogonal since
$\mathbf{r}\!\perp\!\mathbf{e}_\varphi$, and is aligned with the $+\lambda$ direction.
We read $(\sin\lambda,\cos\lambda)$ off the ray as
$x/\cos\varphi,\,z/\cos\varphi$ with $\cos\varphi=\sqrt{\max(1-y^2,\epsilon_{\mathrm{b}})}$
($\epsilon_{\mathrm{b}}\!=\!10^{-7}$, distinct from the LAC clamp $\epsilon\!=\!10^{-3}$ in
Eq.~(5), main); these ratios stay finite because $x,z\!\to\!0$ together at the
poles, so the frame is well-defined at every coarse-grid location (grid token
centers never sample the exact poles, so this clamp is effectively inactive).
We then evaluate the spherical
log-map (Eq.~(3), main) in a stable form with an inner-product clamp
(keeping $\theta\!<\!\pi$) and a norm clamp (removing the $0/0$ pole form), so
the resulting bias is finite for every token pair. Because TPB forms an all-to-all $(N\!\times\!N)$
table on the $32\!\times\!64$ coarse grid, exact antipodal pairs occur; these
are only $O(N)$ of the $O(N^2)$ entries, and the norm clamp assigns them a
bounded, deterministic bias value, so $\boldsymbol{\delta}_{q,k}$ is
geometrically exact away from this sparse set.
\paragraph{Encoding and MLP.} Before Fourier encoding, the offset is
scaled per query by $\cos^{\alpha}\!\varphi_q$ with a single learnable scalar
$\alpha$ initialized at $0$, which lets the bias adapt its latitude
conditioning. The scaled offset is encoded with $8$ geometric frequencies
$(1,2,\dots,2^{7})$ per tangent coordinate as $(\sin,\cos)$ pairs
($32$ dimensions) and mapped by a $32\!\to\!32\!\to\!1$ MLP (GELU,
zero-initialized output; $1{,}089$ parameters); $\alpha$ and this MLP are the
only learned parts of TPB.

\section{Stanford2D3D Test Protocol}
\label{sec:supp-s2d3d}

\noindent\textbf{Frame source.}
Stanford2D3D~\cite{stanford2d3d} provides $1{,}413$ equirectangular panoramas at
$4096\!\times\!2048$ across $6$ indoor areas (area 5 is split into
$5a$/$5b$ as separately scanned wings). We drop one truncated panorama in
area 3, leaving $\mathbf{1{,}412}$ usable panoramas (area 1: 190, area 2:
299, area 3: 84, area 4: 258, area 5a: 143, area 5b: 230, area 6: 208).

\noindent\textbf{Coordinate harmonization.}
The dataset stores camera poses as $3\!\times\!4$ world-to-camera matrices in
a $+y$-down image-frame convention and depth as $16$-bit PNGs with scale
$1/512$\,m. Two normalizations match our runtime ERP convention (Sec.~3, main):
\textit{(i)} a $\mathrm{diag}(1, -1, 1)$ y-flip on each pose's rotation
block, so image-top corresponds to $+y$ (up); \textit{(ii)} the $/512$ depth
scaling to meters.

\noindent\textbf{Pair generation.}
Within each area we enumerate all uni-directional pairs $(i, j)$, $i\!<\!j$,
and compute an EDM-style overlap score:
\[
  \mathrm{ov}(i, j) \;=\;
  \frac{|\{\, p \in V_i \,:\, |d_{i\to j}(p) - d_j(\Pi(p))| \,/\, d_j(\Pi(p)) < 0.1\,\}|}
       {|V_i|},
\]
where $V_i$ is the set of valid-depth pixels of $I_i$, $\Pi$ is the
depth-and-pose-driven ERP reprojection from $I_i$ to $I_j$, and the relative-depth
consistency threshold $0.1$ follows EDM~\cite{edm}. We retain pairs with $\mathrm{ov}(i, j) \in [0.30, 0.80]$.

\noindent\textbf{Resulting pair count.}
\begin{center}
\begin{tabular}{lrrr}
\toprule
Area & Frames & All pairs $(i\!<\!j)$ & Retained ($0.30\!\le\!\mathrm{ov}\!\le\!0.80$) \\
\midrule
area 1  & 190 & 17{,}955 & 1{,}057 \\
area 2  & 299 & 44{,}551 & 3{,}403 \\
area 3  &  84 &  3{,}486 &    233 \\
area 4  & 258 & 33{,}153 & 1{,}178 \\
area 5a & 143 & 10{,}153 &    620 \\
area 5b & 230 & 26{,}335 & 1{,}118 \\
area 6  & 208 & 21{,}528 & 1{,}135 \\
\midrule
Total   & 1{,}412 & 157{,}161 & \textbf{8{,}744} \\
\bottomrule
\end{tabular}
\end{center}
All $8{,}744$ pairs are used for evaluation.

\noindent\textbf{Difference from EDM's protocol.}
EDM~\cite{edm} retains pairs with $\mathrm{ov} > 0.50$ ($3{,}460$ pairs).
Our $[0.30, 0.80]$ overlap band includes lower-covisibility pairs
and excludes near-duplicate views.

\FloatBarrier
\section{Downstream Evaluation}
\label{sec:supp-pose}

\begin{figure}[htbp]
\centering
\includegraphics[width=\textwidth]{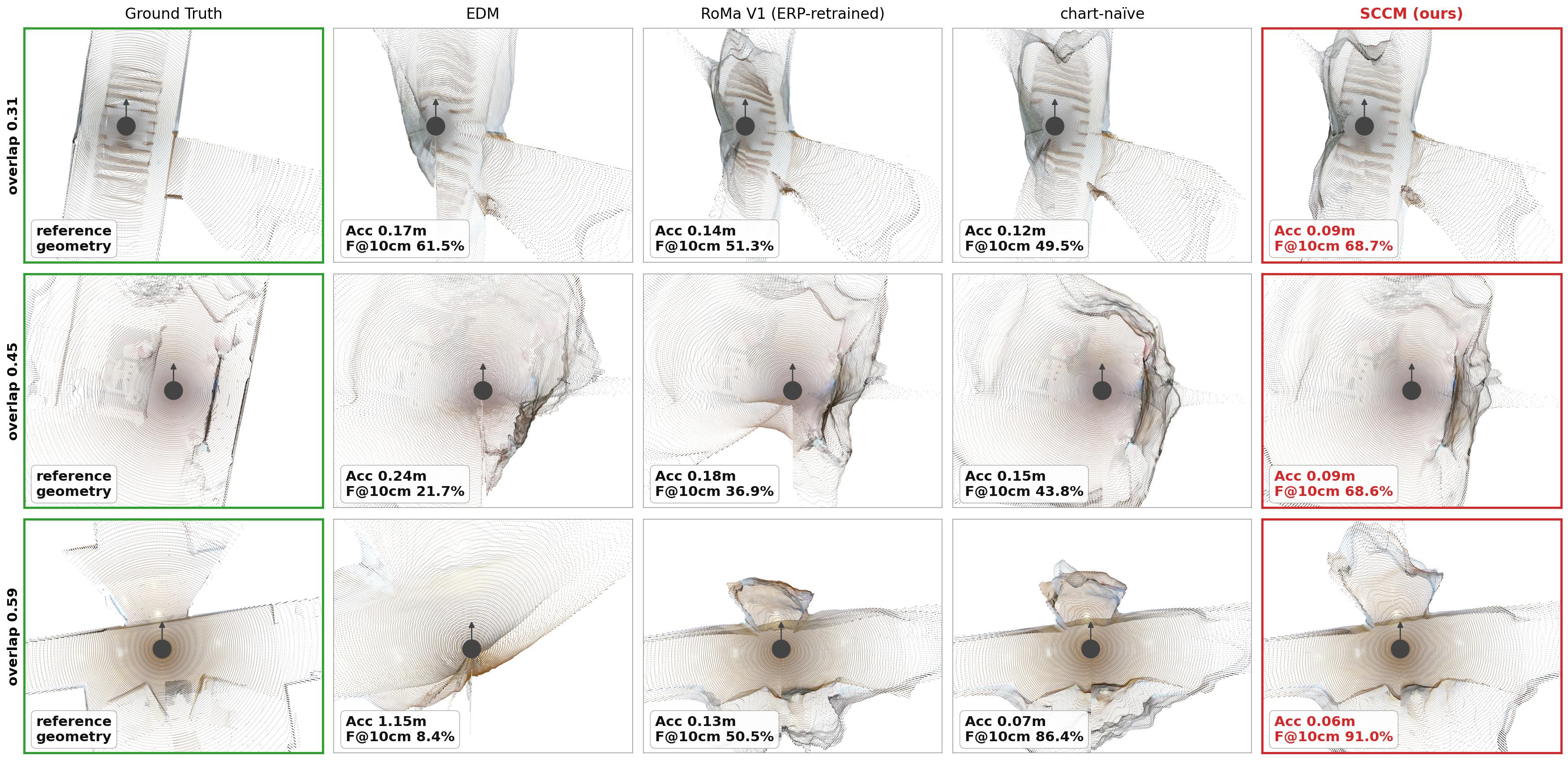}
\caption{\textbf{Top-down reconstruction comparison} on three
representative test pairs (qualitative; aggregate reconstruction metrics
are in Tab.~3, main). Each panel projects the triangulated cloud onto the
X--Z plane (camera $A$ at origin). On these examples the baselines (EDM,
RoMa~V1 (ERP-retrained)) and our chart-naïve scaffold produce distorted room
contours, whereas SCCM's cloud (red) aligns more closely with the GT reference
(green).}
\label{fig:recon_qual}
\end{figure}

The main paper summarizes both downstream tasks (relative pose and 3D
reconstruction) in Sec.~5.4 under one unified protocol. This section provides
the full protocol, the overlap-selected pose subset
(Tab.~\ref{tab:supp-pose-ov}), and top-down reconstructions
(Fig.~\ref{fig:recon_qual}).

\noindent\textbf{Pose.}
For each Matterport3D test pair we sample $2{,}000$ matches by certainty-weighted
multinomial sampling (RoMa~\cite{roma}), unproject them to unit-ray pairs,
and estimate the essential matrix
with the same $8$-point $+$ RANSAC solver for every method, decomposing it to
$(\mathbf{R},\mathbf{t})$ by cheirality; the pose AUC of Tab.~3 (main) is aggregated over the $11{,}574$-pair subset defined under Reconstruction below. To control for pair difficulty, we additionally report an
overlap-selected subset
($\mathrm{ov}\!>\!0.5$, $9{,}268$ pairs; Tab.~\ref{tab:supp-pose-ov}) under the
same pipeline. All methods use this single evaluator, so differences reflect the
matches rather than the estimator.

\begin{table}[!ht]
\centering
\caption{\textbf{Pose on the overlap-selected subset} ($\mathrm{ov}\!>\!0.5$,
$9{,}268$ pairs) under our unified ray-essential evaluator. We report Pose-AUC
using $\max(\text{rotation},\text{translation})$ error.
Best \textbf{bold}, second \underline{underlined}.}
\label{tab:supp-pose-ov}
\setlength{\tabcolsep}{8pt}
\begin{tabular}{l ccc}
\toprule
\textbf{Method} & \textbf{AUC@}$5^\circ$ & \textbf{AUC@}$10^\circ$ & \textbf{AUC@}$20^\circ$ \\
\midrule
EDM~\cite{edm}                       & 11.52 & 24.55 & 38.17 \\
RoMa~V1 (ERP-retrained)~\cite{roma}  & 14.69 & 33.60 & 54.28 \\
chart-naïve                        & \underline{17.56} & \underline{37.68} & \underline{57.97} \\
\textbf{SCCM}                        & \textbf{22.06} & \textbf{42.36} & \textbf{61.17} \\
\bottomrule
\end{tabular}
\end{table}

\noindent\textbf{Reconstruction.}
For each test pair we triangulate stride-$4$ pixels of image~$A$ against their
predicted matches under the GT pose (two-ray intersection, chei\-rality-filtered),
using a shared per-pixel certainty gate $\tau\!=\!0.5$---the same certainty
signal the pose pipeline samples from (RoMa~\cite{roma}, DKM~\cite{dkm}). All
methods are scored on the common $11{,}574$-pair subset where every method
yields $\ge\!200$ confident matches; the triangulated cloud is compared to the GT
depth cloud for accuracy, completeness, and F-score at 5/10/20\,cm. The ungated
variant (no gate) is discussed in Sec.~\ref{sec:supp-recontail}.

\section{Performance Gains over the Chart-Naïve Baseline}
\label{sec:supp-bootstrap}

Tab.~\ref{tab:supp-bootstrap} summarizes the SCCM\,$-$\,chart-naïve
differences on Matterport3D and zero-shot Stanford2D3D.
SCCM improves all five metrics on both datasets, with PCK@$1^\circ$ gains
of $4.5$\,pp and $5.0$\,pp, respectively.
Training variability is assessed in Sec.~\ref{sec:supp-controls}.

\begin{table}[!htb]
\centering
\caption{\textbf{Gains over the chart-naïve baseline} on Matterport3D test
($15{,}682$ pairs) and zero-shot Stanford2D3D ($8{,}744$ pairs).
PCK is fraction-of-pixels ($\uparrow$); MAE/median are in degrees
($\downarrow$). Differences are computed from the displayed values.
Best in \textbf{bold}.}
\label{tab:supp-bootstrap}
{\footnotesize
\setlength{\tabcolsep}{1.5pt}
\resizebox{\columnwidth}{!}{%
\begin{tabular}{llccccc}
\toprule
\textbf{Set} & \textbf{Method} & PCK@$1^\circ\!\uparrow$ & PCK@$3^\circ\!\uparrow$ & PCK@$5^\circ\!\uparrow$ & MAE$\downarrow$ & Med$\downarrow$ \\
\midrule
\multirow{3}{*}{MP3D}
 & chart-naïve & $0.230$ & $0.609$ & $0.769$ & $5.68$ & $2.23$ \\
 & \textbf{SCCM} & $\mathbf{0.275}$ & $\mathbf{0.665}$ & $\mathbf{0.806}$ & $\mathbf{5.36}$ & $\mathbf{1.88}$ \\
 & $\Delta$ & $+0.045$ & $+0.056$ & $+0.037$ & $-0.32$ & $-0.35$ \\
\midrule
\multirow{3}{*}{Stanford2D3D}
 & chart-naïve & $0.179$ & $0.475$ & $0.627$ & $18.86$ & $3.26$ \\
 & \textbf{SCCM} & $\mathbf{0.229}$ & $\mathbf{0.564}$ & $\mathbf{0.697}$ & $\mathbf{17.09}$ & $\mathbf{2.45}$ \\
 & $\Delta$ & $+0.050$ & $+0.089$ & $+0.070$ & $-1.77$ & $-0.81$ \\
\bottomrule
\end{tabular}%
}
}
\end{table}

\section{Sphere-Area-Weighted Metric}
\label{sec:supp-areaweighted}

We recompute every metric with per-pixel weights proportional to
$\cos\varphi$ to account for sphere area.
Table~\ref{tab:supp-areaweighted} shows that the chart-naïve\,$\to$\,SCCM
gain persists on every metric under area weighting, on both Matterport3D
and zero-shot Stanford2D3D.

\begin{table}[!htb]
\centering
\caption{\textbf{Sphere-area-weighted metrics}: each pixel weighted by $\cos\varphi$ vs
the pixel-uniform grid, on Matterport3D test and zero-shot Stanford2D3D. SCCM
leads all baselines (EDM, ERP-retrained RoMa~V1, the chart-naïve scaffold)
on every metric under both pixel-uniform and sphere-area-weighted aggregation,
on both datasets. The pixel-uniform columns reproduce the main-paper tables;
the area-weighted columns re-weight the same test pixels by $\cos\varphi$.}
\label{tab:supp-areaweighted}
{\footnotesize
\setlength{\tabcolsep}{3pt}
\resizebox{\columnwidth}{!}{%
\begin{tabular}{llccccc}
\toprule
 & & PCK@$1^\circ\!\uparrow$ & PCK@$3^\circ\!\uparrow$ & PCK@$5^\circ\!\uparrow$ & MAE$\downarrow$ & Med$\downarrow$ \\
\midrule
\multicolumn{7}{l}{\emph{Matterport3D test}}\\
\multirow{2}{*}{EDM} & pixel-uniform & $0.163$ & $0.400$ & $0.509$ & $16.78$ & $4.79$ \\
 & area-weighted & $0.186$ & $0.419$ & $0.518$ & $17.16$ & $4.55$ \\
\multirow{2}{*}{RoMa~V1 (ERP-retrained)} & pixel-uniform & $0.198$ & $0.538$ & $0.711$ & $6.60$ & $2.76$ \\
 & area-weighted & $0.238$ & $0.565$ & $0.716$ & $6.84$ & $2.44$ \\
\multirow{2}{*}{chart-naïve} & pixel-uniform & $0.230$ & $0.609$ & $0.769$ & $5.68$ & $2.23$ \\
 & area-weighted & $0.264$ & $0.622$ & $0.764$ & $5.98$ & $2.06$ \\
\multirow{2}{*}{\textbf{SCCM}} & pixel-uniform & $\mathbf{0.275}$ & $\mathbf{0.665}$ & $\mathbf{0.806}$ & $\mathbf{5.36}$ & $\mathbf{1.88}$ \\
 & area-weighted & $\mathbf{0.303}$ & $\mathbf{0.667}$ & $\mathbf{0.795}$ & $\mathbf{5.74}$ & $\mathbf{1.78}$ \\
\midrule
\multicolumn{7}{l}{\emph{Zero-shot Stanford2D3D}}\\
\multirow{2}{*}{EDM} & pixel-uniform & $0.104$ & $0.262$ & $0.355$ & $35.70$ & $10.69$ \\
 & area-weighted & $0.136$ & $0.302$ & $0.387$ & $35.11$ & $9.68$ \\
\multirow{2}{*}{RoMa~V1 (ERP-retrained)} & pixel-uniform & $0.167$ & $0.452$ & $0.628$ & $17.80$ & $3.45$ \\
 & area-weighted & $0.215$ & $0.494$ & $0.643$ & $17.27$ & $3.06$ \\
\multirow{2}{*}{chart-naïve} & pixel-uniform & $0.179$ & $0.475$ & $0.627$ & $18.86$ & $3.26$ \\
 & area-weighted & $0.221$ & $0.504$ & $0.638$ & $17.99$ & $2.96$ \\
\multirow{2}{*}{\textbf{SCCM}} & pixel-uniform & $\mathbf{0.229}$ & $\mathbf{0.564}$ & $\mathbf{0.697}$ & $\mathbf{17.09}$ & $\mathbf{2.45}$ \\
 & area-weighted & $\mathbf{0.274}$ & $\mathbf{0.588}$ & $\mathbf{0.710}$ & $\mathbf{16.25}$ & $\mathbf{2.20}$ \\
\bottomrule
\end{tabular}%
}
}
\end{table}

\section{Off-Gravity Behavior Beyond Upright ERP}
\label{sec:supp-tilt}

SCCM assumes upright ERP. To probe this assumption we synthesize off-gravity
captures from the test set: for a fixed pitch rotation $R$ we resample both
panoramas by $R$ and conjugate the relative pose $T'\!=\!GTG^{-1}$
($G=\mathrm{blkdiag}(R,1)$), which preserves the GT correspondences up to
resampling artifacts while tipping the chart off the gravity horizon.
Table~\ref{tab:supp-tilt} reports PCK@$1^\circ$ for SCCM, its chart-naïve
scaffold, EDM, and ERP-retrained RoMa~V1 on the full test split under
$10^\circ$/$20^\circ$/$30^\circ$ pitch. All ERP matchers degrade off-gravity;
SCCM is not tilt-equivariant, but owing to its larger upright margin it retains
the highest \emph{absolute} PCK@$1^\circ$ at every tested tilt. A natively tilt-equivariant matcher remains future work.

\begin{table}[ht]
\centering
\caption{\textbf{Off-gravity tilt robustness} on Matterport3D test
($15{,}682$ pairs). A fixed SO(3) pitch is applied to both views and the
relative pose is conjugated accordingly, so the ground truth remains consistent
up to resampling artifacts. We report PCK@$1^\circ$ under
increasing pitch. All ERP matchers degrade off-gravity; SCCM is not
tilt-equivariant but retains the highest absolute accuracy at every tested
angle. Best per column in \textbf{bold}.}
\label{tab:supp-tilt}
\begin{tabular}{lcccc}
\toprule
Method & upright (resampled) & pitch $10^\circ$ & pitch $20^\circ$ & pitch $30^\circ$ \\
\midrule
EDM~\cite{edm}                      & 0.163 & 0.105 & 0.044 & 0.018 \\
RoMa~V1 (ERP-retrained)~\cite{roma} & 0.198 & 0.096 & 0.042 & 0.022 \\
chart-naïve                         & 0.230 & 0.099 & 0.043 & 0.023 \\
\midrule
SCCM (ours)                         & \textbf{0.273} & \textbf{0.142} & \textbf{0.064} & \textbf{0.034} \\
\bottomrule
\end{tabular}
\end{table}

\section{Rationale for Backbone and Baseline Training Choices}
\label{sec:supp-backbone}

\noindent\textbf{Why not V2.}
We use RoMa~V1's public training recipe for the controlled
$1$\,M-sample protocol. A comparable public recipe for
RoMa~V2~\cite{romav2} was unavailable at the time of this study. Moreover, V2's intertwined positional changes
(normalized-grid RoPE, fixed frequency $\omega = 1$, multi-view Transformer
with alternating attention) interact directly with the attention-side modules of
SPA, so re-applying SPA on V2 would confound ``what the V2 positional changes
provide'' with ``what our sphere-aware prior provides.''

\noindent\textbf{V1 is the appropriate scaffold for controlled ablation.}
On the fixed chart-naïve attention scaffold (cross-attention $+$
dual-softmax, no PE) that replaces V1's GP coarse stage, we add our
yaw-periodic RoPE, tangent-plane bias, and AAC cumulatively; this is the
scaffold used for the cumulative ablation (Tab.~2, main) and the factorial
decomposition (Sec.~\ref{sec:supp-analysis}). V2 is retained as a zero-shot
perspective baseline in Tab.~1 (main), where its released inference
checkpoint suffices.

\noindent\textbf{EDM baseline.}
EDM's~\cite{edm} public release provides inference code and a checkpoint
trained on Matterport3D; the training code, loss, and data pipeline
are not released. Retraining EDM under our $1$\,M-sample protocol would
therefore require re-implementing its full training stack, with the same
reproduction confound noted for V2 above. We instead evaluate EDM from its
released in-domain checkpoint (trained by its authors on the same
Matterport3D data under their own recipe and budget), and separately
evaluate its input-side absolute spherical PE as a mechanism-only control
retrained under the identical protocol (Tab.~2, main;
Sec.~\ref{sec:supp-edm-pe}).

\noindent\textbf{Portability to other matchers.}
SPA and AAC attach to two generic interfaces rather than to RoMa-specific
components: SPA to the pairwise attention logits of the coarse matcher, and AAC
to its per-pixel pre-sigmoid covisibility logits. Our chart-na\"ive scaffold
R1 combines cross-attention, a covisibility head, and dual-softmax, and the
cumulative ablation (Tab.~2, main) validates the modules on this scaffold.
Transfer to other attention-based matchers, including LoFTR~\cite{loftr},
requires support for the positional rotations and additive pairwise bias
used by SPA, and a per-pixel covisibility logit for AAC. GP-based coarse stages (DKM~\cite{dkm}, original
RoMa~V1~\cite{roma}) can accommodate AAC through their covisibility logits,
but SPA additionally requires an attention-based coarse stage, as in R1.

\section{Yaw-Periodic RoPE: Frequency Choice}
\label{sec:supp-rope-bw}

Restricting the longitudinal RoPE frequencies to integers
($\omega_i^{\lambda} \in \mathbb{Z}$, Sec.~4.1~(i) main) is the only choice
consistent with $2\pi$-periodicity at the chart seam: any fractional
$\omega^{\lambda}$ shifts the cross-seam relative phase by a non-multiple of
$2\pi$ and breaks the cyclicity SPA enforces.
At coarse-grid width $W=64$, the longitudinal Nyquist limit is $32$
cycles per $360^\circ$. Our maximum frequency
$\omega_{\max}=d_h/4=16$ ($d_h=64$) is half this limit;
its $22.5^\circ$ period spans four token spacings.
The integer-frequency design outperforms the geometric-frequency
control in Tab.~2 of the main paper.
Only the longitudinal half of each head is restricted; the latitude half
keeps the standard geometric schedule
($\omega_i^{\varphi} = \theta_0^{-2i/(d_h/2)}$, $\theta_0 = 10000$), so fine
latitude discrimination is preserved as in the latitude subspace of standard RoPE.

\section{Chart-na\"ive Coarse Matcher (R1 Backbone)}
\label{sec:supp-chartnaive}

\begin{sloppypar}
The chart-naïve backbone (row R1) keeps RoMa~V1's~\cite{roma} frozen
DINOv2-Large~\cite{dinov2} encoder and the ConvRefiner \emph{architecture}
(the refiner weights, like all non-encoder modules, are trained from scratch),
and replaces \emph{only} V1's Gaussian-process coarse matcher with a standard
attention $+$ dual-softmax coarse stage. It carries \emph{no} sphere-aware component---no yaw-periodic RoPE,
tangent-plane bias, EDM abs.\ PE, or log-area correction---and every ablation
row, the EDM abs.\ PE control, and SCCM is this same backbone augmented with one
or more such terms, which is what makes the comparison controlled.
\end{sloppypar}

\noindent\textbf{Coarse stage.}
Coarse features $\mathbf{x}_A,\mathbf{x}_B\in\mathbb{R}^{N\times d}$
($d\!=\!512$, $N\!=\!H\!\times\!W$ the coarse grid) pass through a cascade
of $n_{\text{blocks}}\!=\!4$ Cross\,$\to$\,Self attention blocks ($8$ heads,
each with a feed-forward sub-layer) using \emph{no} positional
encoding and \emph{no} relative-position bias. Assignment is by dual-softmax
\begin{equation}
\mathbf{P} = \mathrm{softmax}_{\text{row}}(\mathbf{S}/\tau)\,\odot\,
\mathrm{softmax}_{\text{col}}(\mathbf{S}/\tau),\quad
\mathbf{S}=\mathbf{x}_A\mathbf{x}_B^{\!\top},\ \ \tau\!=\!0.1,
\end{equation}
and the soft correspondence (the expected target coordinate under the
row-normalized $\mathbf{P}$) is passed to the RoMa~V1 ConvRefiner.

\noindent\textbf{Covisibility gate.}
A per-query head produces a covisibility logit
$\ell_p=\mathrm{MLP}(\mathbf{x}_p)$ ($128$-d hidden) mapped to a gate
$\mu_p=\sigma(\ell_p)$ that down-weights non-covisible queries. In R1 this is
the bare MLP; AAC (Sec.~4.2, main) adds the single log-area term on top of
this same logit. All components except the frozen encoder are randomly
initialized and trained from scratch under the protocol of
Sec.~\ref{sec:supp-impl}.

\section{EDM Positional Embedding: Re-implementation}
\label{sec:supp-edm-pe}

\noindent\textbf{Scope of this probe.}
The purpose here is to isolate a single design axis---an \emph{absolute,
input-side} sphere-coordinate positional encoding (as in EDM) vs.\ our
\emph{relative, intra-attention} bias (SPA)---under one identical training
protocol, data split, and evaluator. To do so we place EDM's positional
embedding (EDM Eq.~(1) and Eq.~(9)) on the same R1 backbone
(cross-attention $+$ dual-softmax) used by SCCM. We implement EDM's
published PE equations rather than using EDM's full system. Any conclusion drawn here
therefore bears on the \emph{absolute-PE mechanism itself}, not on EDM as
a whole system (which also has a distinct backbone and a geodesic-flow refiner).

\noindent\textbf{Formulation.}
For each coarse-grid token at latitude/longitude $(\varphi,\lambda)$ we
form the 3D unit ray (EDM Eq.~(1))
\begin{equation}
\mathbf{r}(\varphi,\lambda) =
(\sin\lambda\cos\varphi,\; \sin\varphi,\; \cos\lambda\cos\varphi),
\end{equation}
the same unit ray $\mathbf{r}(\varphi,\lambda)$ used by our TPB
(Sec.~4.1, main). The embedding (EDM Eq.~(9)) is
\begin{equation}
\boldsymbol{\chi} = \cos(\mathbf{W}_{\mathrm{PE}}\,\mathbf{r} + \mathbf{b}),
\qquad \mathbf{W}_{\mathrm{PE}}\in\mathbb{R}^{D\times 3},\ \mathbf{b}\in\mathbb{R}^{D},
\end{equation}
with $\mathbf{W}_{\mathrm{PE}}$ a learned $1\times1$ projection ($3\!\to\!D$),
$\mathbf{b}$ a learned bias, and $\cos$ element-wise. $\boldsymbol{\chi}$
is added to \emph{both} token sets \emph{before} the attention cascade
($\mathbf{x}\!\leftarrow\!\mathbf{x}+\boldsymbol{\chi}$,
$\mathbf{y}\!\leftarrow\!\mathbf{y}+\boldsymbol{\chi}$). It is purely
absolute (per-token, independent of the query--key pair): there is no
relative-phase mechanism, which is precisely the property that
distinguishes it from SPA.

\noindent\textbf{Initialization.}
We initialize $\mathbf{W}_{\mathrm{PE}}=\mathbf{0}$ and
$\mathbf{b}=\tfrac{\pi}{2}$, giving $\boldsymbol{\chi}=\mathbf{0}$
at initialization.

\noindent\textbf{Reproducibility.}
We implement only the EDM absolute sphere-coordinate embedding (unit ray
Eq.~(1) $\to$ $\cos$ of a learned $3\!\to\!D$ projection Eq.~(9) $\to$ added to
both token grids before the R1 coarse matcher); all other EDM components are not
used.

\FloatBarrier
\section{AAC/LAC Numerical Sanity Checks}
\label{sec:supp-eps}

\noindent\textbf{Clamp sensitivity.}
The numerical safety clamp $\epsilon=10^{-3}$ in Eq.~(5) (AAC) of the main paper
is active only in an extremely thin numerical pole band
($|\varphi|\!>\!89.94^\circ$, ${\sim}0.06\%$ of latitudes); all remaining pixels see
the exact unclamped $\cos\varphi$. It acts on the non-trainable $\cos\varphi$
buffer, so the learnable $\alpha_{\mathrm{LAC}}$ (entering only as
$\alpha_{\mathrm{LAC}}\!\cdot\!\log[\cdot]$) retains a well-defined gradient
dominated by the unclamped majority. Tab.~\ref{tab:supp-eps} shows that varying
$\epsilon$ over $[10^{-4},10^{-2}]$ leaves every metric unchanged.
\begin{table}[ht]
\centering
\caption{Sensitivity to the numerical safety clamp $\epsilon$ used in
Eq.~(5) of the main paper, on the full Matterport3D \emph{test} split
($15{,}682$ pairs) with the full SCCM model. Varying $\epsilon$ over
$[10^{-4},10^{-2}]$ leaves all metrics unchanged at reported precision,
confirming that the clamp is not driving the result. PCK $\uparrow$; MAE/median in degrees $\downarrow$.}
\label{tab:supp-eps}
\setlength{\tabcolsep}{5pt}
\begin{tabular}{cccccc}
\toprule
$\epsilon$ & PCK@$1^\circ\!\uparrow$ & PCK@$3^\circ\!\uparrow$ & PCK@$5^\circ\!\uparrow$ & MAE$^\circ\!\downarrow$ & Med$^\circ\!\downarrow$ \\
\midrule
$10^{-4}$        & $0.275$ & $0.665$ & $0.806$ & $5.36$ & $1.88$ \\
$10^{-3}$ (used) & $0.275$ & $0.665$ & $0.806$ & $5.36$ & $1.88$ \\
$10^{-2}$        & $0.275$ & $0.665$ & $0.806$ & $5.36$ & $1.88$ \\
\bottomrule
\end{tabular}
\end{table}

\medskip
\noindent\textbf{LAC strength.}
\label{sec:supp-lac-strength}
The learned LAC strengths remain near the analytic area value
($\alpha_{\mathrm{LAC}}^{A}=0.978$, $\alpha_{\mathrm{LAC}}^{B}=1.028$; init
$\alpha=1$). Overriding them to $\alpha=1$ at inference time leaves the metrics
within $10^{-4}$ (Tab.~\ref{tab:supp-lac}), indicating low sensitivity
to the final scalar value at inference.

\begin{table}[ht]
\centering
\caption{\textbf{LAC strength sanity check} on the full Matterport3D
test split. The trained SCCM checkpoint is evaluated with its \emph{learned}
per-view LAC strengths $(\alpha_{\mathrm{LAC}}^{A},\alpha_{\mathrm{LAC}}^{B})$ and
with $\alpha$ \emph{fixed to $1$ at inference time}. The two settings yield
metrics within $10^{-4}$ of each other.}
\label{tab:supp-lac}
\setlength{\tabcolsep}{4pt}
\begin{tabular}{lccccc}
\toprule
\textbf{Model} & $\alpha_{\mathrm{LAC}}^{A}/\alpha_{\mathrm{LAC}}^{B}$
 & PCK@$1^\circ\!\uparrow$ & PCK@$3^\circ\!\uparrow$
 & PCK@$5^\circ\!\uparrow$ & MAE$^\circ\!\downarrow$ \\
\midrule
SCCM (learned $\alpha$)        & $0.978/1.028$ & $0.275$ & $0.665$ & $0.806$ & $5.36$ \\
SCCM ($\alpha\!=\!1$, infer.)  & $1.000/1.000$ & $0.275$ & $0.665$ & $0.806$ & $5.36$ \\
\bottomrule
\end{tabular}
\end{table}

\FloatBarrier
\section{Computational Complexity Analysis}
\label{sec:supp-complexity}

Tab.~\ref{tab:complexity} reports trainable parameters and FLOPs. SCCM adds only
a parameter-free yaw-periodic RoPE, an $\approx\!1.1$K-parameter TPB MLP, and one
LAC scalar per covisibility side; the TPB bias costs ${\approx}8.9$\,GFLOPs
(${\approx}1.2\%$) when recomputed every forward and nothing when cached.
Tab.~\ref{tab:latency-cache} reports latency: uncached TPB recomputation is
memory-bound, but caching the content-independent TPB table reduces the deployed
overhead to ${\sim}5\%$. We detail the FLOPs bound and latency protocol below.

\begin{table}[ht]
\centering
\caption{\textbf{Computational footprint} relative to the R1 chart-naïve
baseline: trainable-parameter and FLOP additions at the medium ERP resolution
($448\!\times\!896$). Inference latency is reported separately in Tab.~\ref{tab:latency-cache}.}
\label{tab:complexity}
\scriptsize
\setlength{\tabcolsep}{5pt}
\renewcommand{\arraystretch}{0.95}
\begin{tabular*}{0.9\columnwidth}{@{\extracolsep{\fill}}lrr}
\toprule
\textbf{Configuration} & \textbf{Trainable params} & \textbf{GFLOPs} \\
\midrule
(R1) chart-naïve (cross-attn$+$DS) & $137.17$\,M & $751.15$ \\
$+$ SPA (RoPE + TPB)      & $+\!\approx\!1.1$\,k & $\approx\!9.0$ ($\approx\!0.1$ cached) \\
$+$ AAC (LAC)             & $+\!2$            & $\approx\!0.0$ \\
\midrule
\textbf{SCCM total} & $137.17$\,M & $\approx\!760.2$ ($\approx\!751.3$ cached) \\
\textbf{Increase vs.\ baseline} & $<\!0.002\%$ & $\approx\!1.2\%$ ($\approx\!0.01\%$ cached) \\
\bottomrule
\end{tabular*}
\end{table}

\subsection{FLOPs}
\label{sec:supp-flops}

The chart-naïve baseline forward pass measures $\mathbf{751.15}$~GFLOPs at
the medium ERP resolution $448 \times 896$ (\texttt{thop}, RTX 4090). The SCCM
additions are bounded analytically. RoPE, a per-channel rotation of queries and keys in each attention
layer, costs about $0.1$ GFLOPs in total, and AAC is an $N$-element pointwise
term with negligible cost. TPB evaluates
its $32\!\to\!32\!\to\!1$ MLP on all $N^2 = 2048^2$ token pairs,
${\approx}4.4$\,GMACs ${\approx}8.9$\,GFLOPs, i.e.\ ${\approx}1.2\%$ of the
$751$\,GFLOPs baseline when the bias is recomputed every forward, and $0$ when
the content-independent bias is cached (Sec.~\ref{sec:supp-latency}). The
wall-clock overhead of the uncached path (Tab.~\ref{tab:latency-cache})
exceeds this compute share because it is memory-bound (an $N\!\times\!N\!\times\!32$
intermediate) with kernel-launch overhead; caching eliminates it, leaving a
deployed ${\sim}5\%$.

\subsection{Inference Latency Protocol}
\label{sec:supp-latency}

The latency values in Tab.~\ref{tab:latency-cache} are wall-clock times on an RTX 4090,
single image pair at the medium ERP resolution, mean over $100$
forward passes after a $10$-iteration warm-up. The pairwise tangent log-map
buffer (Sec.~4.1~(ii), main) is precomputed and excluded from the
per-forward timing, consistent with deployment.

\noindent\textbf{TPB bias caching.}
The TPB bias is a fixed grid-geometry prior at inference: it depends only on
the token positions, not on image content, so it can be precomputed once and
reused for every pair; the cache is behavior-preserving and disabled during training. Tab.~\ref{tab:latency-cache} reports a
back-to-back A/B on a single RTX~4090: recomputing the bias each forward costs
$+20.9\%$, whereas caching it reduces SCCM's overhead to $+5.0\%$.
We use the cached configuration for deployment.

\begin{table}[ht]
\centering
\caption{\textbf{TPB bias caching latency.} Wall-clock latency on a single
RTX~4090 at medium ERP resolution, averaged over $100$ timed forwards after
warm-up. The cached variant precomputes the content-independent TPB bias for the
fixed ERP grid and reuses it at inference time.}
\label{tab:latency-cache}
\setlength{\tabcolsep}{8pt}
\begin{tabular}{lcc}
\toprule
\textbf{Method} & \textbf{Latency (ms)} & \textbf{Overhead} \\
\midrule
chart-naïve            & $115.76$ & --- \\
SCCM (TPB uncached)    & $139.95$ & $+20.90\%$ \\
SCCM (TPB cached)      & $121.60$ & $\mathbf{+5.04\%}$ \\
\bottomrule
\end{tabular}
\end{table}

\FloatBarrier
\section{Module Contributions and Interaction}
\label{sec:supp-analysis}

\begin{figure}[ht]
\centering
\includegraphics[width=0.90\textwidth]{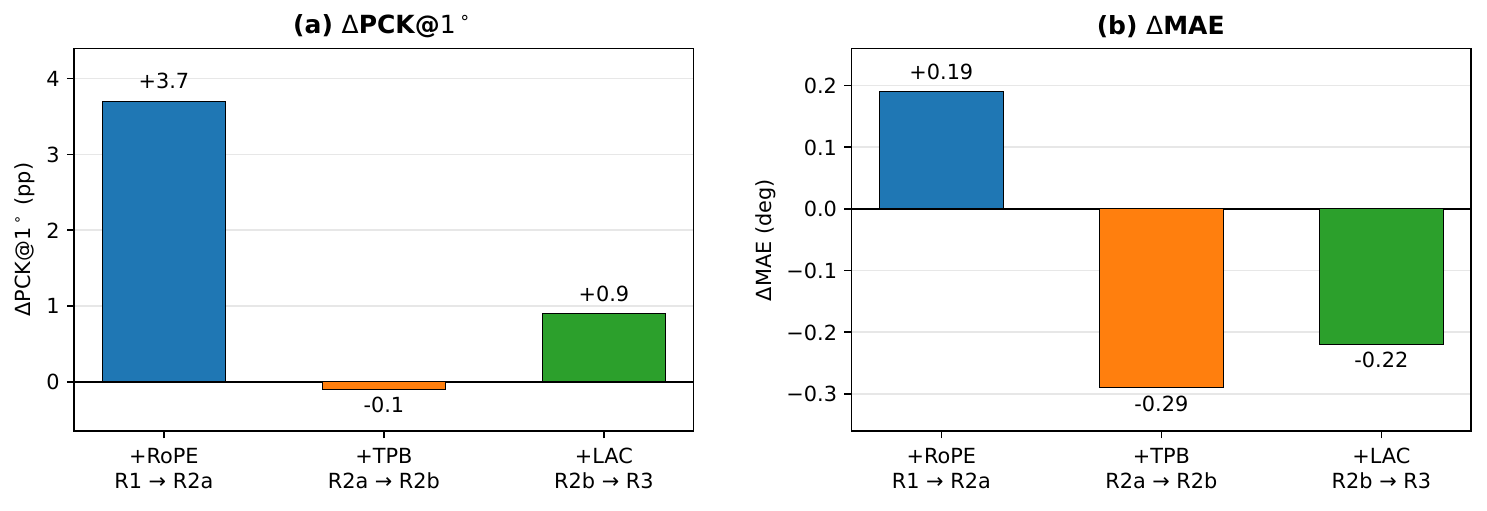}
\caption{\textbf{Marginal changes along R1$\to$R2a$\to$R2b$\to$R3}
from Tab.~2 (main): blue $+$yaw-periodic RoPE (R2a$-$R1), orange $+$TPB
(R2b$-$R2a), and green $+$LAC (R3$-$R2b).
(a)~$\Delta$PCK@$1^\circ$ (pp); (b)~$\Delta$MAE (deg).
Positive $\Delta$PCK and negative $\Delta$MAE indicate improvement.}
\label{fig:supp-analysis}
\end{figure}

Fig.~\ref{fig:supp-analysis} summarizes the marginal changes along the
R1$\to$R2a$\to$R2b$\to$R3 ablation in Tab.~2 (main).
Yaw-periodic RoPE provides the largest PCK@$1^\circ$ gain ($+3.7$\,pp),
with a $+0.19^\circ$ change in MAE.
TPB reduces MAE by $0.29^\circ$ with a $-0.1$\,pp change in PCK@$1^\circ$.
Adding LAC to full SPA improves PCK@$1^\circ$ by $0.9$\,pp and reduces MAE
by $0.22^\circ$. These are cumulative marginal effects; the factorial
analysis below separates the TPB$\times$LAC interaction.

\noindent\textbf{TPB error-tail quantiles.}
To verify that TPB acts on the angular-error tail rather than strict threshold
precision, we compare pixel-level angular-error quantiles of R2a and R2b on the
full Matterport3D test set (Tab.~\ref{tab:supp-tpb-quantiles}). TPB leaves
PCK@$1^\circ$ and the median (P50) essentially unchanged, but reduces the
high-error quantiles (P90/P95/P99 by $0.41^\circ$/$1.80^\circ$/$4.96^\circ$).
This confirms the interpretation in Tab.~2 (main) that TPB calibrates the
pairwise metric and suppresses large-error tails rather than improving
near-threshold precision.

\begin{table}[ht]
\centering
\caption{\textbf{TPB error-tail quantiles} on Matterport3D test (full split,
${\sim}3.0$B valid pixels). R2a is yaw-periodic RoPE only; R2b adds TPB (full
SPA). Angular-error quantiles P$k$ are in degrees; in the $\Delta$ row,
PCK@$1^\circ$ is in percentage points and all other columns in degrees. Differences are computed from the displayed values. TPB leaves strict precision (PCK@$1^\circ$) and the median
(P50) nearly unchanged but lowers the high-error quantiles, confirming its role
as tail calibration.}
\label{tab:supp-tpb-quantiles}
\setlength{\tabcolsep}{4pt}
\footnotesize
\begin{tabular}{lccccccc}
\toprule
Row & PCK@$1^\circ\,\uparrow$ & MAE$\,\downarrow$ & P50$\,\downarrow$ & P75$\,\downarrow$ & P90$\,\downarrow$ & P95$\,\downarrow$ & P99$\,\downarrow$ \\
\midrule
R2a (RoPE)     & 0.267 & 5.87 & 1.93 & 4.18 & 9.29 & 19.49 & 94.23 \\
R2b (full SPA) & 0.266 & 5.58 & 1.95 & 4.13 & 8.88 & 17.69 & 89.27 \\
$\Delta$       & $-0.1$ & $-0.29$ & $+0.02$ & $-0.05$ & $-0.41$ & $-1.80$ & $-4.96$ \\
\bottomrule
\end{tabular}
\end{table}

\noindent\textbf{Factorial decomposition: the two priors interact.}
The cumulative ladder of Tab.~2 (main) credits each gain to the last added
term. To separate the roles of TPB and LAC we train the missing combination
(yaw-periodic RoPE $+$ LAC, no TPB) under the identical protocol and
evaluate all four combinations, completing a
$2\times2$ factorial over the R2a base (Tab.~\ref{tab:supp-factorial}).
Individually, TPB and LAC each change PCK@$1^\circ$ by $-0.1$\,pp;
together they improve it by $+0.8$\,pp. The resulting interaction is
$+1.0$\,pp. Both priors reduce MAE individually
($-0.29^\circ$ and $-0.34^\circ$), with conditional reductions of
$-0.17^\circ$ and $-0.22^\circ$, respectively.
The combination improves strict PCK@$1^\circ$, consistent with
their complementary roles: LAC adjusts the matching mass retained by
the area-aware gate, while TPB adjusts its distribution under the
spherical metric. This distribution determines the expected target
coordinate (Sec.~\ref{sec:supp-chartnaive}).

\begin{table}[ht]
\centering
\caption{\textbf{TPB$\times$LAC factorial} on Matterport3D test ($15{,}682$
pairs): all four combinations over the R2a (yaw-periodic RoPE) base under the
shared protocol of Tab.~\ref{tab:supp-protocol}.
Effects and the interaction are computed from the displayed values.}
\label{tab:supp-factorial}
\footnotesize
\setlength{\tabcolsep}{3pt}
\begin{tabular}{lcc cccc}
\toprule
\textbf{Cell} & TPB & LAC & PCK@$1^\circ\!\uparrow$ & PCK@$3^\circ\!\uparrow$ & PCK@$5^\circ\!\uparrow$ & MAE$\downarrow$ \\
\midrule
R2a (RoPE)        & \ding{55} & \ding{55} & $0.267$ & $0.653$ & $0.796$ & $5.87$ \\
R2b ($+$TPB)      & \ding{51} & \ding{55} & $0.266$ & $0.652$ & $0.799$ & $5.58$ \\
$+$LAC (no TPB)   & \ding{55} & \ding{51} & $0.266$ & $0.658$ & $0.802$ & $5.53$ \\
R3 (SCCM)         & \ding{51} & \ding{51} & $\mathbf{0.275}$ & $\mathbf{0.665}$ & $\mathbf{0.806}$ & $\mathbf{5.36}$ \\
\midrule
\multicolumn{3}{l}{\textbf{Effect}} & \multicolumn{2}{c}{$\Delta$PCK@$1^\circ$ (pp)} & \multicolumn{2}{c}{$\Delta$MAE (deg)} \\
\midrule
\multicolumn{3}{l}{TPB alone (R2b$-$R2a)}      & \multicolumn{2}{c}{$-0.1$} & \multicolumn{2}{c}{$-0.29$} \\
\multicolumn{3}{l}{LAC alone}                  & \multicolumn{2}{c}{$-0.1$} & \multicolumn{2}{c}{$-0.34$} \\
\multicolumn{3}{l}{TPB given LAC}              & \multicolumn{2}{c}{$+0.9$} & \multicolumn{2}{c}{$-0.17$} \\
\multicolumn{3}{l}{LAC given TPB (R3$-$R2b)}   & \multicolumn{2}{c}{$+0.9$} & \multicolumn{2}{c}{$-0.22$} \\
\multicolumn{3}{l}{\textbf{Interaction}}       & \multicolumn{2}{c}{$\mathbf{+1.0}$} & \multicolumn{2}{c}{$+0.12$} \\
\bottomrule
\end{tabular}
\end{table}

\section{Analysis of Remaining Errors}
\label{sec:supp-failures}

\begin{figure}[t]
\centering
\includegraphics[width=0.8\columnwidth]{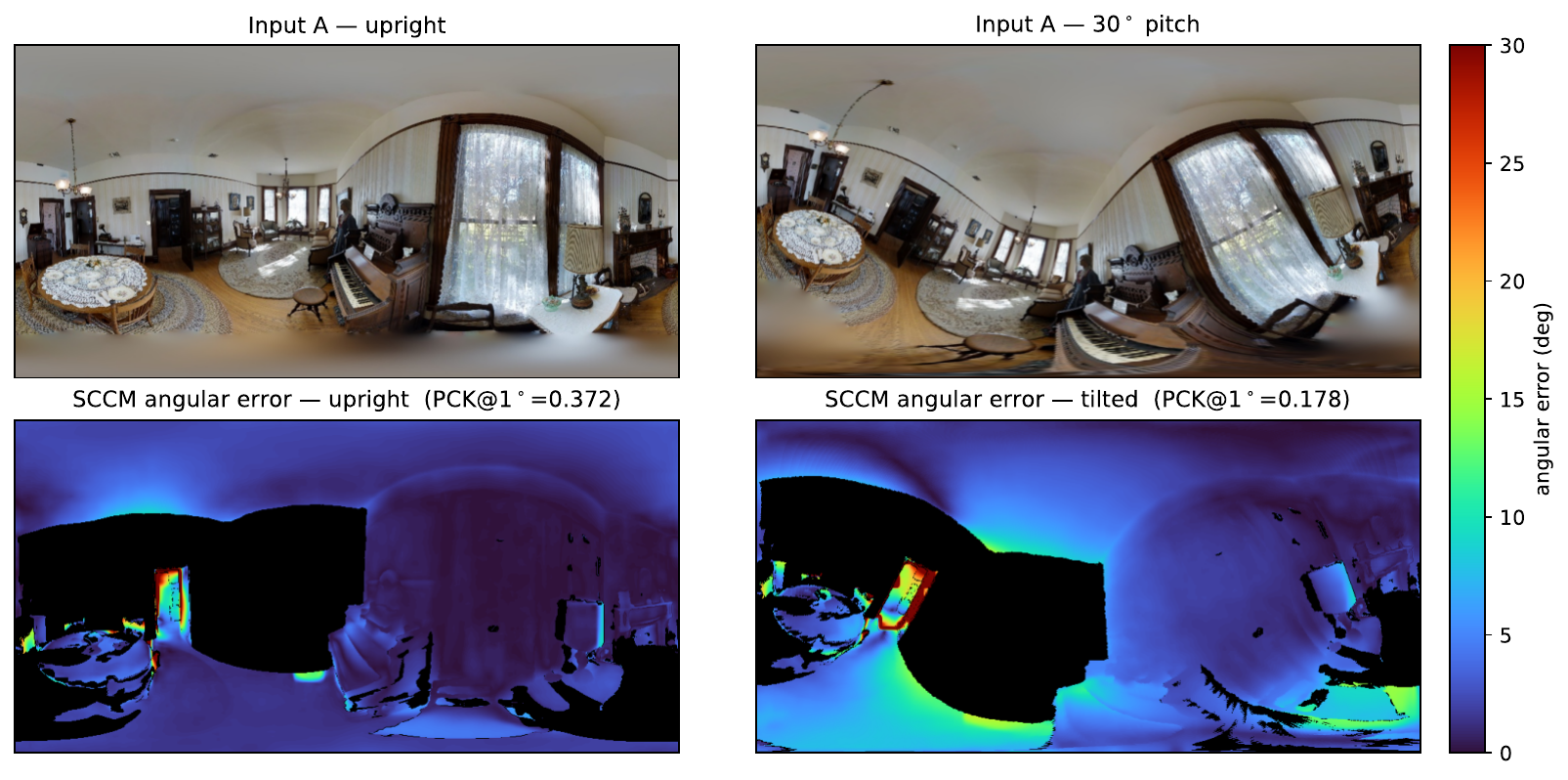}
\caption{\textbf{SCCM's off-gravity failure mode}, shown on a single
Matterport3D test pair (the \emph{same} pair throughout).
\emph{Top}: input view~$A$ upright vs.\ under a synthetic $30^\circ$
off-gravity pitch (Sec.~\ref{sec:supp-tilt}), which tips the ERP chart off the
gravity horizon. \emph{Bottom}: SCCM's per-pixel angular error for the warp
$A\!\to\!B$ (black~$=$~non-covisible; colorbar $0^\circ$ to $\ge\!30^\circ$).
Upright, SCCM attains PCK@$1^\circ\!=\!0.37$ on this pair (mostly low error);
under tilt the gravity-aligned prior no longer matches the tilted chart and
accuracy drops markedly (PCK@$1^\circ\!=\!0.18$), with error concentrated where
the chart distortion is largest. Holding the pair fixed isolates tilt---not
scene content---as the cause; full-set tilt averages are in
Tab.~\ref{tab:supp-tilt}. This is a geometry-specific failure outside the intended
upright-ERP setting.}
\label{fig:supp-failure}
\end{figure}

SCCM's residual errors fall into two groups.
\textbf{(a)~Geometry-specific failures} occur outside the intended upright-ERP
setting: a large camera pitch/roll breaks the gravity-aligned prior and accuracy
drops sharply (PCK@$1^\circ$ $0.273\!\to\!0.034$ at a $30^\circ$ tilt;
Fig.~\ref{fig:supp-failure}, Sec.~\ref{sec:supp-tilt}). Within upright ERP, the
polar safety clamps are inactive because no coarse-grid token lies in their
pole band (Secs.~\ref{sec:supp-impl} and~\ref{sec:supp-eps}); the separate
handling of exact antipodal token pairs is described in Sec.~\ref{sec:supp-impl}.
\textbf{(b)~Generic dense-matching failures} remain in texture-less,
reflective/HDR, low-overlap, and depth-discontinuity regions; these are shared by
perspective and ERP matchers alike (cf.\ RoMa~\cite{roma}, EDM~\cite{edm},
LoFTR~\cite{loftr}), are driven by visual ambiguity rather than chart geometry,
and are orthogonal to the spherical priors---requiring upstream changes (a
stronger encoder, photometric-invariant features, occlusion handling).

\subsection{Certainty-gated vs.\ ungated reconstruction}
\label{sec:supp-recontail}

The main reconstruction protocol uses a shared certainty gate
$\tau\!=\!0.5$ for all methods. Without this gate, SCCM's mean accuracy
is affected by a rare low-certainty tail, while its median accuracy
and majority-pair performance remain better than chart-na\"ive.

\section{Refinement-Stage Diagnosis}
\label{sec:supp-refstage}

To locate where the mechanism-isolated SCCM gain enters the pipeline, we measure
PCK@$1^\circ$ after the coarse stage and after each refinement stage of the
shared cascade, on the full Matterport3D test split
(${\sim}3.0$B valid pixels). Tab.~\ref{tab:supp-refstage} shows that
SCCM's advantage over the chart-naïve scaffold is already present at the
coarse-anchor stage ($+4.1$\,pp) and remains essentially unchanged through
refinement ($+4.5$\,pp at the final dense output), with small non-monotonic
fluctuation across intermediate stages ($+4.1$ to $+5.4$\,pp). The architecturally unchanged RoMa~V1 refiner
provides a similar coarse-to-final lift for both models ($+10.8$\,pp for
chart-naïve, $+11.2$\,pp for SCCM), preserving and slightly amplifying the
coarse-stage advantage rather than creating it. This supports our interpretation
that the ERP-specific gain enters primarily through coarse-anchor formation
rather than being created by the refiner.

\begin{table}[!htb]
\centering
\caption{\textbf{Refinement-stage PCK@$1^\circ$ diagnostic} on the full
Matterport3D test split. PCK@$1^\circ$ after the
coarse stage and after each refinement stage. SCCM's advantage is present before
refinement and preserved through the architecturally unchanged RoMa~V1 refiner; the similar
coarse-to-final lift for both models indicates that the gain enters primarily
through coarse-anchor formation rather than being created by the refiner.}
\label{tab:supp-refstage}
\setlength{\tabcolsep}{8pt}
\begin{tabular}{lccc}
\toprule
\textbf{Stage} & \textbf{chart-naïve} & \textbf{SCCM} & \textbf{$\Delta$ (pp)} \\
\midrule
coarse anchor ($1/16$) & 0.122 & 0.163 & $+4.1$ \\
refiner stage ($1/8$)  & 0.133 & 0.187 & $+5.4$ \\
refiner stage ($1/4$)  & 0.188 & 0.234 & $+4.6$ \\
refiner stage ($1/2$)  & 0.219 & 0.263 & $+4.4$ \\
final dense            & \textbf{0.230} & \textbf{0.275} & $+4.5$ \\
\bottomrule
\end{tabular}
\end{table}

\FloatBarrier


\section{Outdoor Evaluation on Holo360D}
\label{sec:supp-holo360d}

This section details the outdoor experiment summarized in Tab.~1(c) of the main
paper. Holo360D~\cite{holo360d} is an in-the-wild dataset captured with a
handheld LiDAR scanner coupled to a 360$^\circ$ camera along continuous
trajectories. Its outdoor sequences are far deeper and sparser in supervision
than Matterport3D (median scene depth $4.3$\,m vs.\ $2.0$\,m; $54\%$ vs.\ $95\%$
of ERP pixels carry valid depth). The median relative tilt between paired
views is $12^\circ$, with $79\%$ of pairs exceeding $5^\circ$.

\paragraph{Protocol.} The three retrained matchers of Tab.~1 (main)---the ERP-retrained RoMa~V1 (V1-GP), the chart-na\"ive scaffold (R1),
and SCCM---are initialized from their Matterport3D checkpoints and trained on Holo360D with a shared training budget
($25$k optimizer steps, learning-rate scale $0.1$, same augmentation and loss),
on scene-disjoint splits of $5$ training, $3$ validation, and $4$ test scenes,
and score $8{,}000$ held-out test pairs ($2{,}000$ per test scene; overlap
$0.3$--$0.8$) with the angular evaluator of the main paper.
For SCCM's Holo360D training, the RoPE rotation angles are multiplied by
$\gamma\!\sim\!U[0,1]$ at each forward pass; evaluation uses $\gamma\!=\!1$.
This stochastic regularization is specific to Holo360D training.

\begin{table}[!htb]
\centering
\caption{\textbf{Outdoor training on Holo360D} ($8{,}000$ test pairs). All
three matchers start from their Matterport3D checkpoints and are trained
with the settings above; SCCM leads on every metric, including the two error
statistics. Best in \textbf{bold}.}
\label{tab:supp-holo360d}
{\footnotesize
\resizebox{\columnwidth}{!}{%
\begin{tabular}{lccccccc}
\toprule
Method & PCK@$0.35^\circ\!\uparrow$ & PCK@$0.5^\circ\!\uparrow$ & PCK@$1^\circ\!\uparrow$ & PCK@$3^\circ\!\uparrow$ & PCK@$5^\circ\!\uparrow$ & MAE$^\circ\!\downarrow$ & Med.$^\circ\!\downarrow$ \\
\midrule
RoMa~V1 (ERP-retrained) & 0.123 & 0.182 & 0.322 & 0.592 & 0.726 & 5.66 & 2.12 \\
chart-na\"ive (R1)      & 0.119 & 0.180 & 0.331 & 0.626 & 0.761 & 5.02 & 1.93 \\
\textbf{SCCM}           & \textbf{0.137} & \textbf{0.202} & \textbf{0.357} & \textbf{0.645} & \textbf{0.776} & \textbf{4.94} & \textbf{1.76} \\
\bottomrule
\end{tabular}}}
\end{table}

\paragraph{Result.} SCCM leads on all seven metrics
(Tab.~\ref{tab:supp-holo360d}), including the two stricter thresholds not shown
in the main table ($+1.8$\,pp at $0.35^\circ$), with a $+2.6$\,pp margin
over R1 at $1^\circ$.

\paragraph{Zero-shot transfer.} Applying the same Holo360D-trained models
\emph{unchanged} to a second outdoor corpus (Mapillary Metropolis~\cite{metropolis}; $2{,}484$
street-level pairs resampled to match the rotation distribution of the Holo360D
test set) tests transfer without any adaptation. SCCM retains its lead at tight
thresholds ($+1.9$\,pp PCK@$0.35^\circ$, $+0.5$\,pp PCK@$1^\circ$ over R1) but
not at $3^\circ$--$5^\circ$ or in mean angular error ($10.08^\circ$ vs.\
$9.39^\circ$ for R1), i.e.\ its zero-shot advantage is confined to precision.

\section{Additional Zero-Shot Baselines on ERP}
\label{sec:supp-zeroshot}

Tab.~\ref{tab:supp-zeroshot} extends Tab.~1 of the main paper with four further
published matchers, evaluated zero-shot on ERP under the same angular evaluator,
each released checkpoint run at its own operating resolution with the ground
truth on that grid (as for EDM in Sec.~\ref{sec:supp-controls}). All four fall well below EDM, the strongest
released ERP-native baseline, indicating limited zero-shot ERP performance of these checkpoints.

\begin{table}[!htb]
\centering
\caption{\textbf{Additional zero-shot baselines} (PCK@$1^\circ$). $^\dagger$Sparse
matchers scored on their own matches only. EDM from Tab.~1 for reference.}
\label{tab:supp-zeroshot}
{\footnotesize
\begin{tabular}{lcc}
\toprule
Method & Matterport3D & Stanford2D3D \\
\midrule
DKM~\cite{dkm} (CVPR'23)                & 0.035 & 0.026 \\
MASt3R$^\dagger$~\cite{mast3r} (ECCV'24) & 0.057 & 0.028 \\
LoFTR$^\dagger$~\cite{loftr} (CVPR'21)   & 0.105 & 0.058 \\
VGGT~\cite{vggt} (CVPR'25)              & 0.000 & 0.000 \\
\midrule
EDM~\cite{edm} (reference)              & 0.163 & 0.104 \\
\bottomrule
\end{tabular}}
\end{table}

\section{Additional Controls: Seeds, Resolution, and Pose Solver}
\label{sec:supp-controls}

\paragraph{Training-seed variance.} The main ablation trains each configuration
once. To bound training-side variance we retrained the two endpoints of the
ablation, chart-na\"ive (R1) and SCCM, with two additional seeds each under the
identical protocol and re-scored the full Matterport3D test split
(Tab.~\ref{tab:supp-seeds}). PCK@$1^\circ$ ranges from $0.272$ to $0.283$
for SCCM and from $0.223$ to $0.230$ for R1.
The per-configuration sample standard deviation is $0.4$--$0.6$\,pp, whereas
the $+0.9$\,pp LAC step of Tab.~2 is only about twice this scale and, being
measured from single runs, should be read as indicative on the training side.

\begin{table}[!htb]
\centering
\caption{\textbf{Multi-seed retraining} of the ablation endpoints (Matterport3D
test, PCK@$1^\circ$). Results from three training runs per model;
run 1 is the paper result. Standard deviations are computed within each model.}
\label{tab:supp-seeds}
{\footnotesize
\begin{tabular}{lcccc}
\toprule
Configuration & run 1 (paper) & run 2 & run 3 & s.d.\ (pp) \\
\midrule
chart-na\"ive (R1) & 0.230 & 0.230 & 0.223 & 0.4 \\
\textbf{SCCM} & \textbf{0.275} & \textbf{0.283} & \textbf{0.272} & 0.6 \\
\bottomrule
\end{tabular}}
\end{table}

\paragraph{EDM resolution control.} EDM~\cite{edm} is evaluated at its native
training resolution ($320\!\times\!640$); its ground truth and metrics are
computed on that grid. Because no re-scaling of SCCM to EDM's resolution is
canonical, we bracket it: (i)~down-sampling SCCM's input to $320\!\times\!640$
and back (matching input information only) gives PCK@$1^\circ$ $0.273$;
(ii)~matching input, ground truth, \emph{and} scored pixels to EDM's grid (the
same $1.47$\,B evaluated pixels) gives $0.279$. Both straddle the reported
$0.275$: re-scaling moves SCCM by $-0.2$/$+0.4$\,pp while
its gap over EDM ($0.163$) stays at $11.0$--$11.6$\,pp.

\paragraph{Pose solver.} We compare the shared ray-space 8-point$+$RANSAC
estimator with Robust~360-8PA~\cite{solarte360} on fixed matches and inliers.
The common $\ge\!200$-confident-matches filter over the four methods yields
$15{,}112$ pairs (Tab.~\ref{tab:supp-solver}). AUC@$5^\circ$ changes by
at most $0.01$\,pp, with the ranking unchanged.

\begin{table}[!htb]
\centering
\caption{\textbf{Solver swap} on $15{,}112$ Matterport3D test pairs: Pose
AUC@$5^\circ$ with our 8-point$+$RANSAC estimator vs.\ Robust~360-8PA on
identical matches and inliers.}
\label{tab:supp-solver}
{\footnotesize
\begin{tabular}{lcc}
\toprule
Method & 8-pt$+$RANSAC & Robust 360-8PA \\
\midrule
EDM~\cite{edm}           & 8.28  & 8.29  \\
RoMa~V1 (ERP-retrained)  & 11.69 & 11.68 \\
chart-na\"ive (R1)       & 14.17 & 14.16 \\
\textbf{SCCM}            & \textbf{17.96} & \textbf{17.96} \\
\bottomrule
\end{tabular}}
\end{table}

\FloatBarrier
\bibliographystyle{splncs04}
\bibliography{main}